\documentclass[10pt,twocolumn,letterpaper]{article}

\usepackage{cvpr}
\usepackage{times}
\usepackage{epsfig}
\usepackage{graphicx}
\usepackage{amsmath}
\usepackage{amssymb}
\usepackage{caption}
\usepackage{subfig}
\usepackage[font={small}]{caption}

\usepackage[pagebackref=true,breaklinks=true,letterpaper=true,colorlinks,bookmarks=false]{hyperref}

\cvprfinalcopy 

\def\cvprPaperID{385} 

\ifcvprfinal\fi
\begin{document}

\newcommand{\norm}[1]{\lVert#1\rVert}


\title{Image-to-Image Translation with Conditional Adversarial Networks}

\vspace{-1mm}
\author{Phillip Isola
\and
Jun-Yan Zhu
\and
Tinghui Zhou
\and
Alexei A. Efros
\and
\vspace{-1mm}
\\
Berkeley AI Research (BAIR) Laboratory, UC Berkeley\\
{\tt\small \{isola,junyanz,tinghuiz,efros\}@eecs.berkeley.edu}
}

\twocolumn[{%
\renewcommand\twocolumn[1][]{#1}%
\vspace{-1em}
\maketitle
\vspace{-1em}
\begin{center}
    \centering
    \vspace{-0.3in}
    \includegraphics[width=\linewidth]{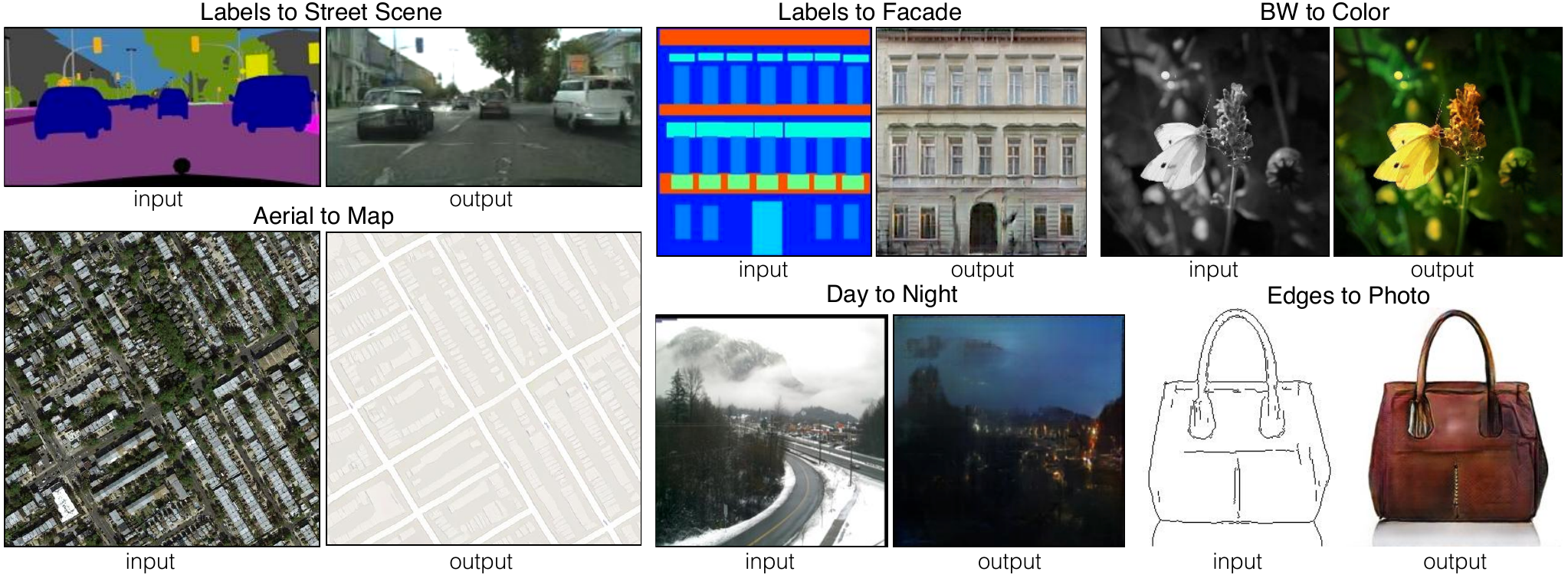}
    \captionof{figure}{Many problems in image processing, graphics, and vision involve translating an input image into a corresponding output image. These problems are often treated with application-specific algorithms, even though the setting is always the same: map pixels to pixels. Conditional adversarial nets are a general-purpose solution that appears to work well on a wide variety of these problems. Here we show results of the method on several. In each case we use the same architecture and objective, and simply train on different data.}
    \label{teaser}
\end{center}%
}]

\vspace{-1mm}
\begin{abstract}
\vspace{-1mm}
We investigate conditional adversarial networks as a general-purpose solution to image-to-image translation problems. These networks not only learn the mapping from input image to output image, but also learn a loss function to train this mapping. This makes it possible to apply the same generic approach to problems that traditionally would require very different loss formulations. We demonstrate that this approach is effective at synthesizing photos from label maps, reconstructing objects from edge maps, and colorizing images, among other tasks. Indeed, since the release of the {\tt pix2pix} software associated with this paper, a large number of internet users (many of them artists) have posted their own experiments with our system, further demonstrating its wide applicability and ease of adoption without the need for parameter tweaking.
As a community, we no longer hand-engineer our mapping functions, and this work suggests we can achieve reasonable results without hand-engineering our loss functions either.

\end{abstract}


\vspace{-1mm}
\section{Introduction}
\vspace{-1mm}
Many problems in image processing, computer graphics, and computer vision can be posed as ``translating'' an input image into a corresponding output image. Just as a concept may be expressed in either English or French, a scene may be rendered as an RGB image, a gradient field, an edge map, a semantic label map, etc. In analogy to automatic language translation, we define automatic \emph{image-to-image translation} as the task of translating one possible representation of a scene into another, given sufficient training data (see Figure \ref{teaser}).
Traditionally, each of these tasks has been tackled with separate, special-purpose machinery (e.g., ~\cite{efros2001image, hertzmann2001image, fergus2006removing, buades2005non, chen2009sketch2photo, shih2013data, laffont2014transient, long2015fully, eigen2015predicting, xie2015holistically, zhang2016colorful}), despite the fact that the setting is always the same: predict pixels from pixels. Our goal in this paper is to develop a common framework for all these problems.

The community has already taken significant steps in this direction, with convolutional neural nets (CNNs) becoming the common workhorse behind a wide variety of image prediction problems. CNNs learn to minimize a loss function -- an objective that scores the quality of results -- and although the learning process is automatic, a lot of manual effort still goes into designing effective losses. In other words, we still have to tell the CNN what we wish it to minimize. But, just like King Midas, we must be careful what we wish for! If we take a naive approach and ask the CNN to minimize the Euclidean distance between predicted and ground truth pixels, it will tend to produce blurry results \cite{pathak2016context, zhang2016colorful}. This is because Euclidean distance is minimized by averaging all plausible outputs, which causes blurring. Coming up with loss functions that force the CNN to do what we really want -- e.g., output sharp, realistic images -- is an open problem and generally requires expert knowledge.

It would be highly desirable if we could instead specify only a high-level goal, like ``make the output indistinguishable from reality", and then automatically learn a loss function appropriate for satisfying this goal. Fortunately, this is exactly what is done by the recently proposed Generative Adversarial Networks (GANs) \cite{goodfellow2014generative, denton2015deep, radford2015unsupervised, salimans2016improved, zhao2016energy}. GANs learn a loss that tries to classify if the output image is real or fake, while simultaneously training a generative model to minimize this loss. Blurry images will not be tolerated since they look obviously fake. Because GANs learn a loss that adapts to the data, they can be applied to a multitude of tasks that traditionally would require very different kinds of loss functions.

In this paper, we explore GANs in the conditional setting. Just as GANs learn a generative model of data, conditional GANs (cGANs) learn a conditional generative model \cite{goodfellow2014generative}. This makes cGANs suitable for image-to-image translation tasks, where we condition on an input image and generate a corresponding output image.

GANs have been vigorously studied in the last two years and many of the techniques we explore in this paper have been previously proposed. Nonetheless, earlier papers have focused on specific applications, and it has remained unclear how effective image-conditional GANs can be as a general-purpose solution for image-to-image translation. Our primary contribution is to demonstrate that on a wide variety of problems, conditional GANs produce reasonable results. Our second contribution is to present a simple framework sufficient to achieve good results, and to analyze the effects of several important architectural choices. Code is available at \texttt{https://github.com/phillipi/pix2pix}.

\section{Related work}

{\bf Structured losses for image modeling} Image-to-image translation problems are often formulated as per-pixel classification or regression (e.g., ~\cite{long2015fully,xie2015holistically,iizuka2016let,larsson2016learning,zhang2016colorful}). These formulations treat the output space as ``unstructured" in the sense that each output pixel is considered conditionally independent from all others given the input image. Conditional GANs instead learn a \emph{structured loss}. Structured losses penalize the joint configuration of the output. A large body of literature has considered losses of this kind, with methods including conditional random fields~\cite{chen14semantic}, the SSIM metric~\cite{wang2004image}, feature matching~\cite{dosovitskiy2016generating}, nonparametric losses~\cite{li2016combining}, the convolutional pseudo-prior~\cite{xie2015convolutional}, and losses based on matching covariance statistics~\cite{johnson2016perceptual}. The conditional GAN is different in that the loss is learned, and can, in theory, penalize any possible structure that differs between output and target.

\begin{figure}[t]
 \centering
 \includegraphics[width=1.0\hsize]{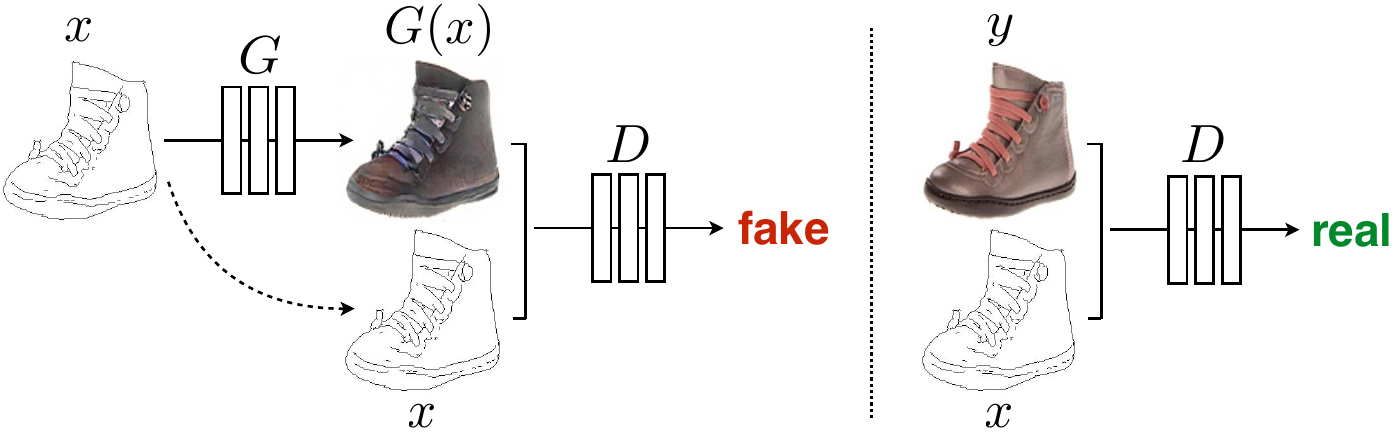}
 \vspace{-0.2in}
  \caption{Training a conditional GAN to map edges$\rightarrow$photo. The discriminator, $D$, learns to classify between fake (synthesized by the generator) and real \{edge, photo\} tuples. The generator, $G$, learns to fool the discriminator. Unlike an unconditional GAN, both the generator and discriminator observe the input edge map.}
 \label{cGAN_method}
 \vspace{-0.2in}
\end{figure}

{\bf Conditional GANs}
We are not the first to apply GANs in the conditional setting. Prior and concurrent works have conditioned GANs on discrete labels~\cite{mirza2014conditional,gauthier2014conditional,denton2015deep}, text~\cite{reed2016generative}, and, indeed, images. The image-conditional models have tackled image prediction from a normal map~\cite{wang2016generative}, future frame prediction~\cite{mathieu2015deep}, product photo generation~\cite{yoo2016pixel}, and image generation from sparse annotations~\cite{karacan2016learning,reed2016learning} (c.f. \cite{reed2016generating} for an autoregressive approach to the same problem). Several other papers have also used GANs for image-to-image mappings, but only applied the GAN unconditionally, relying on other terms (such as L2 regression) to force the output to be conditioned on the input. These papers have achieved impressive results on inpainting~\cite{pathak2016context}, future state prediction~\cite{zhou2016learning}, image manipulation guided by user constraints~\cite{zhu2016generative}, style transfer~\cite{li2016precomputed}, and superresolution \cite{ledig2016photo}. Each of the methods was tailored for a specific application. Our framework differs in that nothing is application-specific. This makes our setup considerably simpler than most others.

Our method also differs from the prior works in several architectural choices for the generator and discriminator. Unlike past work, for our generator we use a ``U-Net"-based architecture~\cite{ronneberger2015u}, and for our discriminator we use a convolutional ``PatchGAN" classifier, which only penalizes structure at the scale of image patches. A similar PatchGAN architecture was previously proposed in~\cite{li2016precomputed} to capture local style statistics. Here we show that this approach is effective on a wider range of problems, and we investigate the effect of changing the patch size.

\section{Method}

GANs are generative models that learn a mapping from random noise vector $z$ to output image $y$, $G: z\rightarrow y$ \cite{goodfellow2014generative}. In contrast, conditional GANs learn a mapping from observed image $x$ and random noise vector $z$, to $y$, $G: \{x,z\}\rightarrow y$. The generator $G$ is trained to produce outputs that cannot be distinguished from ``real" images by an adversarially trained discriminator, $D$, which is trained to do as well as possible at detecting the generator's ``fakes". This training procedure is diagrammed in Figure \ref{cGAN_method}.

\subsection{Objective}

The objective of a conditional GAN can be expressed as
\begin{align}
    \mathcal{L}_{cGAN}(G,D) = &\mathbb{E}_{x,y}[\log D(x,y)] + \nonumber \\
                 &\mathbb{E}_{x,z}[\log (1-D(x,G(x,z))],\label{cGAN_equation}
\end{align}
where $G$ tries to minimize this objective against an adversarial $D$ that tries to maximize it, i.e. $G^*  = \arg\min_G \max_D \mathcal{L}_{cGAN}(G,D)$.

To test the importance of conditioning the discriminator, we also compare to an unconditional variant in which the discriminator does not observe $x$:
\begin{align}
    \mathcal{L}_{GAN}(G,D) = &\mathbb{E}_{y}[\log D(y)] + \nonumber \\
                 &\mathbb{E}_{x,z}[\log (1-D(G(x,z))].\label{GAN_equation}
\end{align}
Previous approaches have found it beneficial to mix the GAN objective with a more traditional loss, such as L2 distance \cite{pathak2016context}. The discriminator's job remains unchanged, but the generator is tasked to not only fool the discriminator but also to be near the ground truth output in an L2 sense. We also explore this option, using L1 distance rather than L2 as L1 encourages less blurring:
\begin{align}
    \mathcal{L}_{L1}(G) = \mathbb{E}_{x,y,z}[\norm{y-G(x,z)}_1].\label{L1_equation}
\end{align}
Our final objective is
\begin{align}
    G^*  = \arg\min_G\max_D \mathcal{L}_{cGAN}(G,D) + \lambda \mathcal{L}_{L1}(G).\label{full_objective}
\end{align}

Without $z$, the net could still learn a mapping from $x$ to $y$, but would produce deterministic outputs, and therefore fail to match any distribution other than a delta function. Past conditional GANs have acknowledged this and provided Gaussian noise $z$ as an input to the generator, in addition to $x$ (e.g., \cite{wang2016generative}). In initial experiments, we did not find this strategy effective -- the generator simply learned to ignore the noise -- which is consistent with Mathieu et al.~\cite{mathieu2015deep}. Instead, for our final models, we provide noise only in the form of dropout, applied on several layers of our generator at both training and test time. Despite the dropout noise, we observe only minor stochasticity in the output of our nets. Designing conditional GANs that produce highly stochastic output, and thereby capture the full entropy of the conditional distributions they model, is an important question left open by the present work.

\subsection{Network architectures}

We adapt our generator and discriminator architectures from those in \cite{radford2015unsupervised}. Both generator and discriminator use modules of the form convolution-BatchNorm-ReLu \cite{ioffe2015batch}. Details of the architecture are provided in the supplemental materials online, with key features discussed below.

\begin{figure}[t]
 \centering
 \includegraphics[width=1.0\hsize]{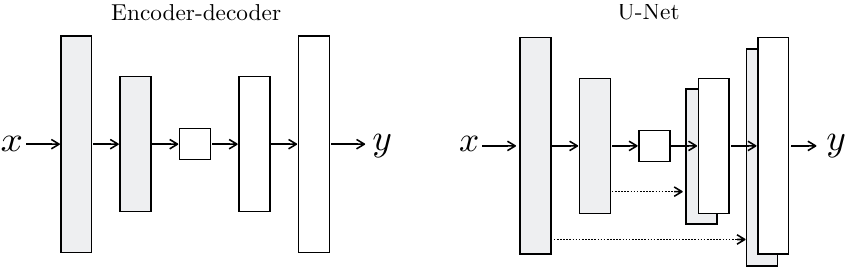}
  \caption{Two choices for the architecture of the generator. The ``U-Net" \cite{ronneberger2015u} is an encoder-decoder with skip connections between mirrored layers in the encoder and decoder stacks.}
 \label{generator_architecture}
\end{figure}

\subsubsection{Generator with skips}
A defining feature of image-to-image translation problems is that they map a high resolution input grid to a high resolution output grid. In addition, for the problems we consider, the input and output differ in surface appearance, but both are renderings of the same underlying structure. Therefore, structure in the input is roughly aligned with structure in the output. We design the generator architecture around these considerations.

Many previous solutions \cite{pathak2016context, wang2016generative, johnson2016perceptual, zhou2016learning, yoo2016pixel} to problems in this area have used an encoder-decoder network \cite{hinton2006reducing}. In such a network, the input is passed through a series of layers that progressively downsample, until a bottleneck layer, at which point the process is reversed. Such a network requires that all information flow pass through all the layers, including the bottleneck. For many image translation problems, there is a great deal of low-level information shared between the input and output, and it would be desirable to shuttle this information directly across the net. For example, in the case of image colorization, the input and output share the location of prominent edges.

To give the generator a means to circumvent the bottleneck for information like this, we add skip connections, following the general shape of a ``U-Net" \cite{ronneberger2015u}. Specifically, we add skip connections between each layer $i$ and layer $n-i$, where $n$ is the total number of layers. Each skip connection simply concatenates all channels at layer $i$ with those at layer $n-i$.

\subsubsection{Markovian discriminator (PatchGAN)}
It is well known that the L2 loss -- and L1, see Figure \ref{cityscapes_loss_variations_qualitative} -- produces blurry results on image generation problems \cite{larsen2015autoencoding}. Although these losses fail to encourage high-frequency crispness, in many cases they nonetheless accurately capture the low frequencies. For problems where this is the case, we do not need an entirely new framework to enforce correctness at the low frequencies. L1 will already do.

This motivates restricting the GAN discriminator to only model high-frequency structure, relying on an L1 term to force low-frequency correctness (Eqn. \ref{full_objective}). In order to model high-frequencies, it is sufficient to restrict our attention to the structure in local image patches. Therefore, we design a discriminator architecture -- which we term a \emph{Patch}GAN -- that only penalizes structure at the scale of patches. This discriminator tries to classify if each $N \times N$ patch in an image is real or fake. We run this discriminator convolutionally across the image, averaging all responses to provide the ultimate output of $D$.

In Section \ref{patch_variations_section}, we demonstrate that $N$ can be much smaller than the full size of the image and still produce high quality results. This is advantageous because a smaller PatchGAN has fewer parameters, runs faster, and can be applied to arbitrarily large images.

Such a discriminator effectively models the image as a Markov random field, assuming independence between pixels separated by more than a patch diameter. This connection was previously explored in \cite{li2016precomputed}, and is also the common assumption in models of texture \cite{efros1999texture,gatys2015texture} and style \cite{efros2001image,hertzmann2001image,gatys2015neural,li2016combining}. Therefore, our PatchGAN can be understood as a form of texture/style loss.

\subsection{Optimization and inference}

To optimize our networks, we follow the standard approach from \cite{goodfellow2014generative}: we alternate between one gradient descent step on $D$, then one step on $G$. As suggested in the original GAN paper, rather than training $G$ to minimize $\log(1-D(x,G(x,z))$, we instead train to maximize $\log D(x,G(x,z))$~\cite{goodfellow2014generative}. In addition, we divide the objective by $2$ while optimizing $D$, which slows down the rate at which $D$ learns relative to $G$. We use minibatch SGD and apply the Adam solver \cite{kingma2014adam}, with a learning rate of $0.0002$, and momentum parameters $\beta_1=0.5$, $\beta_2=0.999$.

At inference time, we run the generator net in exactly the same manner as during the training phase. This differs from the usual protocol in that we apply dropout at test time, and we apply batch normalization \cite{ioffe2015batch} using the statistics of the test batch, rather than aggregated statistics of the training batch. This approach to batch normalization, when the batch size is set to 1, has been termed ``instance normalization" and has been demonstrated to be effective at image generation tasks \cite{ulyanov2016instance}. In our experiments, we use batch sizes between 1 and 10 depending on the experiment.

\section{Experiments}

To explore the generality of conditional GANs, we test the method on a variety of tasks and datasets, including both graphics tasks, like photo generation, and vision tasks, like semantic segmentation:

\begin{itemize}
\itemsep-0.05in
\item \emph{Semantic labels$\leftrightarrow$photo}, trained on the Cityscapes dataset \cite{Cordts2016Cityscapes}.
\item \emph{Architectural labels$\rightarrow$photo}, trained on CMP Facades \cite{Tylecek13}.
\item \emph{Map$\leftrightarrow$aerial photo}, trained on data scraped from Google Maps.
\item \emph{BW$\rightarrow$color photos}, trained on \cite{russakovsky2015imagenet}.
\item \emph{Edges$\rightarrow$photo}, trained on data from \cite{zhu2016generative} and \cite{finegrained_shoes}; binary edges generated using the HED edge detector \cite{xie2015holistically} plus postprocessing.
\item \emph{Sketch$\rightarrow$photo}: tests edges$\rightarrow$photo models on human-drawn sketches from \cite{eitz2012humans}.
\item \emph{Day$\rightarrow$night}, trained on \cite{laffont2014transient}.
\item \emph{Thermal$\rightarrow$color photos}, trained on data from \cite{hwang2015multispectral}.
\item \emph{Photo with missing pixels$\rightarrow$inpainted photo}, trained on Paris StreetView from \cite{doersch2012makes}.
\end{itemize}

\begin{figure}[h]
\begin{center}
\bgroup 
 \def\arraystretch{0.2} 
 \setlength\tabcolsep{0.2pt}
\begin{tabular}{cccc}
 & Classification & Ours & \\
 L2 \cite{zhang2016colorful} & (rebal.) \cite{zhang2016colorful} & (L1 + cGAN) & Ground truth \\
\includegraphics[width=0.25\linewidth]{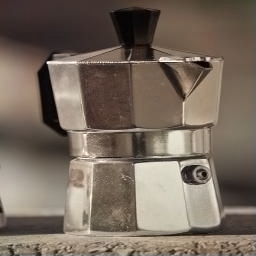} &
\includegraphics[width=0.25\linewidth]{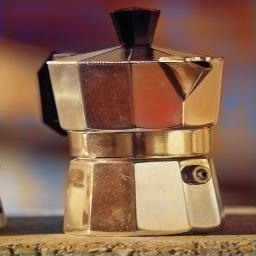} &
\includegraphics[width=0.25\linewidth]{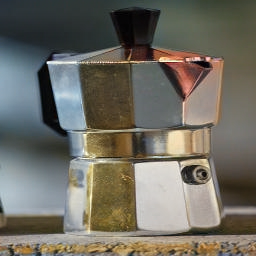} &
\includegraphics[width=0.25\linewidth]{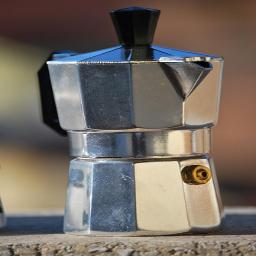} \\ 
\includegraphics[width=0.25\linewidth]{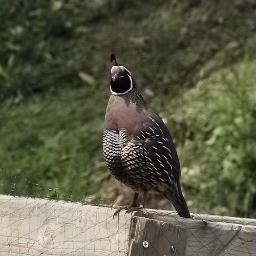} &
\includegraphics[width=0.25\linewidth]{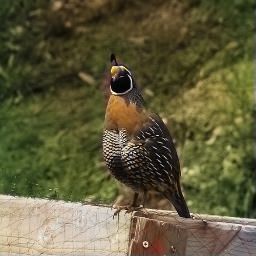} &
\includegraphics[width=0.25\linewidth]{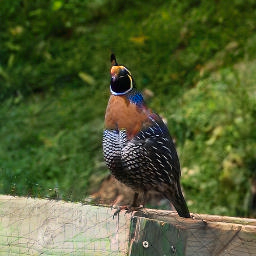} &
\includegraphics[width=0.25\linewidth]{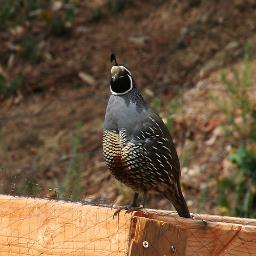} \\ 
\includegraphics[width=0.25\linewidth]{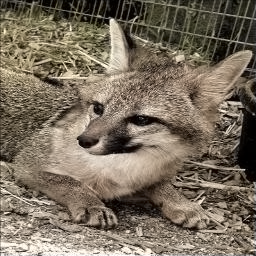} &
\includegraphics[width=0.25\linewidth]{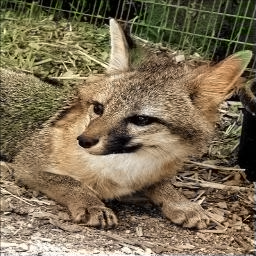} &
\includegraphics[width=0.25\linewidth]{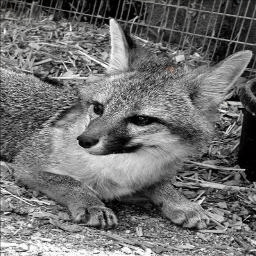} &
\includegraphics[width=0.25\linewidth]{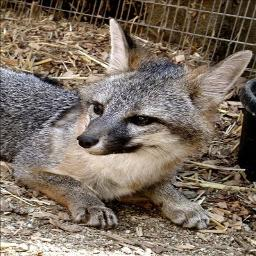} 

\end{tabular} \egroup 
\end{center}
\vspace{-0.1in}
\caption{Colorization results of conditional GANs versus the L2 regression from \cite{zhang2016colorful} and the full method (classification with rebalancing) from \cite{zhou2016learning}. The cGANs can produce compelling colorizations (first two rows), but have a common failure mode of producing a grayscale or desaturated result (last row).}
\vspace{-0.1in}
\label{colorization_res}
\end{figure}

\begin{figure*}
\begin{center}
\bgroup 
 \def\arraystretch{0.2} 
 \setlength\tabcolsep{0.2pt}
\begin{tabular}{cccccc}
Input & Ground truth & Output & Input & Ground truth & Output \\ 
\includegraphics[width=0.16666666666667\linewidth]{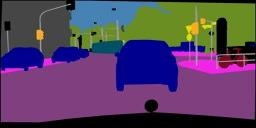} &
\includegraphics[width=0.16666666666667\linewidth]{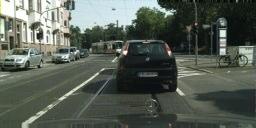} &
\includegraphics[width=0.16666666666667\linewidth]{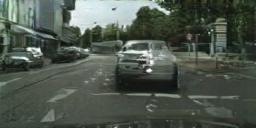} \hspace{0.025in} &
\includegraphics[width=0.16666666666667\linewidth]{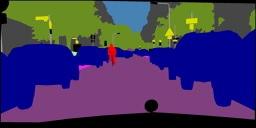} &
\includegraphics[width=0.16666666666667\linewidth]{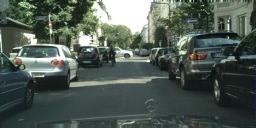} &
\includegraphics[width=0.16666666666667\linewidth]{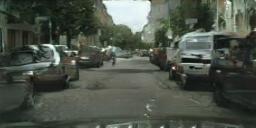} \\ 
\includegraphics[width=0.16666666666667\linewidth]{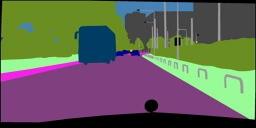} &
\includegraphics[width=0.16666666666667\linewidth]{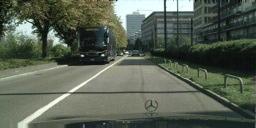} &
\includegraphics[width=0.16666666666667\linewidth]{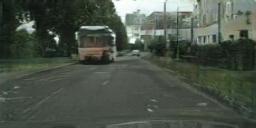} \hspace{0.025in} &
\includegraphics[width=0.16666666666667\linewidth]{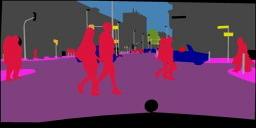} &
\includegraphics[width=0.16666666666667\linewidth]{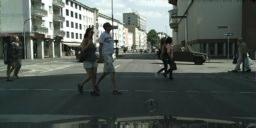} &
\includegraphics[width=0.16666666666667\linewidth]{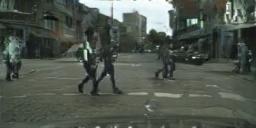} \\ 
\includegraphics[width=0.16666666666667\linewidth]{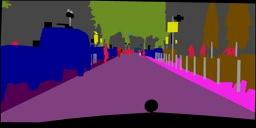} &
\includegraphics[width=0.16666666666667\linewidth]{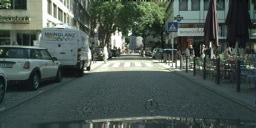} &
\includegraphics[width=0.16666666666667\linewidth]{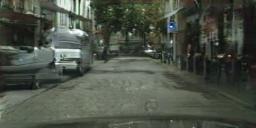} \hspace{0.025in} &
\includegraphics[width=0.16666666666667\linewidth]{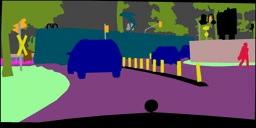} &
\includegraphics[width=0.16666666666667\linewidth]{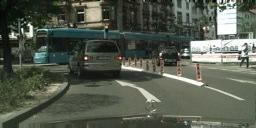} &
\includegraphics[width=0.16666666666667\linewidth]{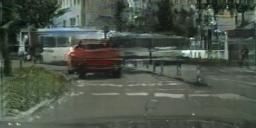} \\ 
\includegraphics[width=0.16666666666667\linewidth]{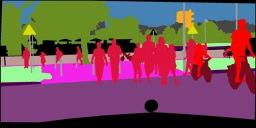} &
\includegraphics[width=0.16666666666667\linewidth]{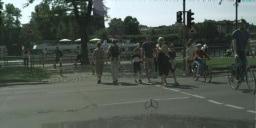} &
\includegraphics[width=0.16666666666667\linewidth]{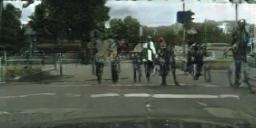} \hspace{0.025in} &
\includegraphics[width=0.16666666666667\linewidth]{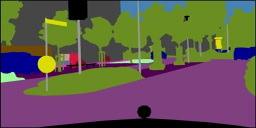} &
\includegraphics[width=0.16666666666667\linewidth]{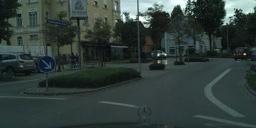} &
\includegraphics[width=0.16666666666667\linewidth]{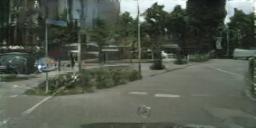}

\end{tabular} \egroup 
\end{center}
\caption{Example results of our method on Cityscapes labels$\rightarrow$photo, compared to ground truth.}
\label{cityscapes_lotsofresults}
\end{figure*}

\begin{figure*}
\begin{center}
\bgroup 
 \def\arraystretch{0.2} 
 \setlength\tabcolsep{0.2pt}
\begin{tabular}{cccccc}
Input & Ground truth & Output & Input & Ground truth & Output \\ 
\includegraphics[width=0.16666666666667\linewidth]{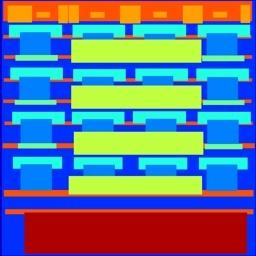} &
\includegraphics[width=0.16666666666667\linewidth]{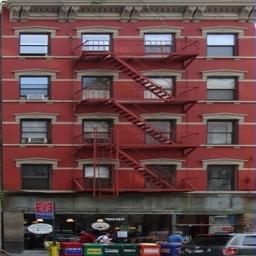} &
\includegraphics[width=0.16666666666667\linewidth]{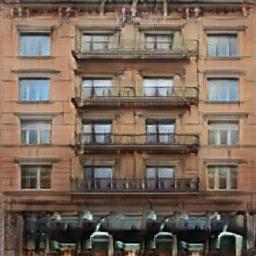} \hspace{0.025in} &
\includegraphics[width=0.16666666666667\linewidth]{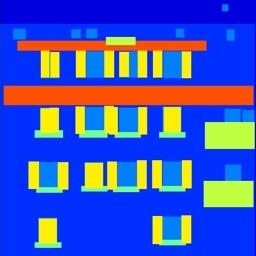} &
\includegraphics[width=0.16666666666667\linewidth]{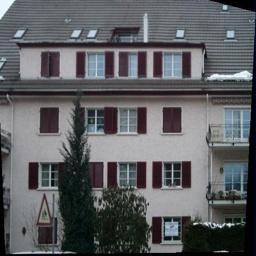} &
\includegraphics[width=0.16666666666667\linewidth]{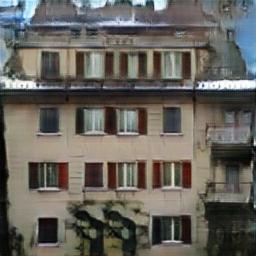} \\ 
\includegraphics[width=0.16666666666667\linewidth]{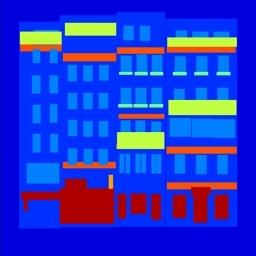} &
\includegraphics[width=0.16666666666667\linewidth]{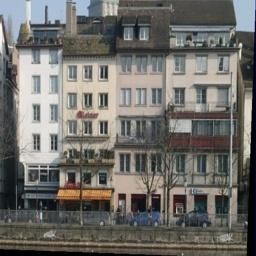} &
\includegraphics[width=0.16666666666667\linewidth]{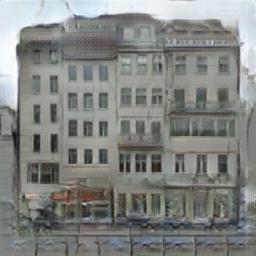} \hspace{0.025in} &
\includegraphics[width=0.16666666666667\linewidth]{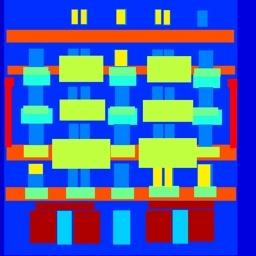} &
\includegraphics[width=0.16666666666667\linewidth]{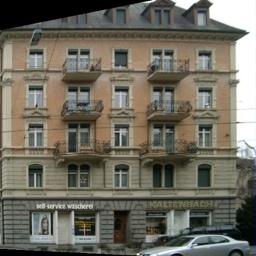} &
\includegraphics[width=0.16666666666667\linewidth]{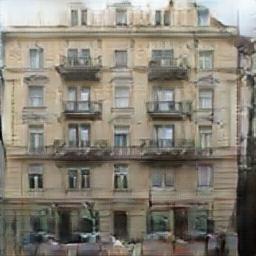} \\ 
\includegraphics[width=0.16666666666667\linewidth]{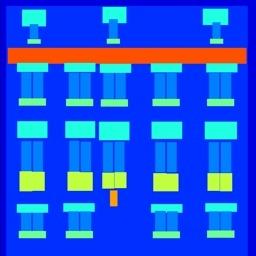} &
\includegraphics[width=0.16666666666667\linewidth]{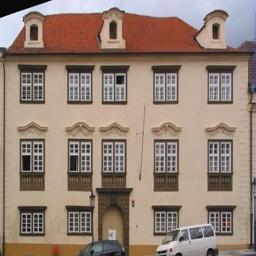} &
\includegraphics[width=0.16666666666667\linewidth]{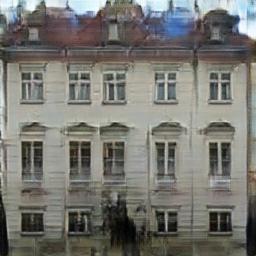} \hspace{0.025in} &
\includegraphics[width=0.16666666666667\linewidth]{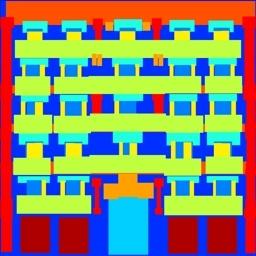} &
\includegraphics[width=0.16666666666667\linewidth]{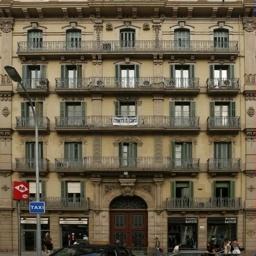} &
\includegraphics[width=0.16666666666667\linewidth]{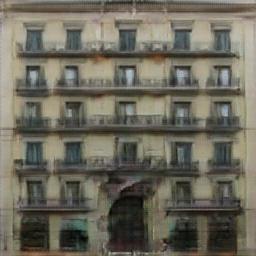} \\ 
\includegraphics[width=0.16666666666667\linewidth]{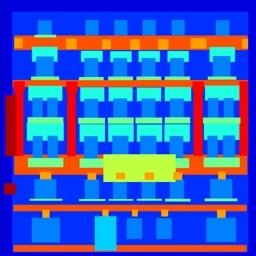} &
\includegraphics[width=0.16666666666667\linewidth]{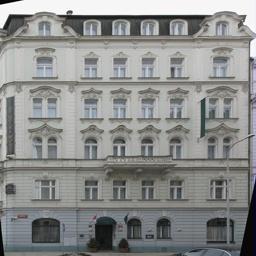} &
\includegraphics[width=0.16666666666667\linewidth]{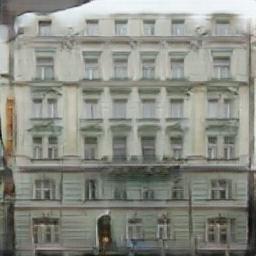} \hspace{0.025in} &
\includegraphics[width=0.16666666666667\linewidth]{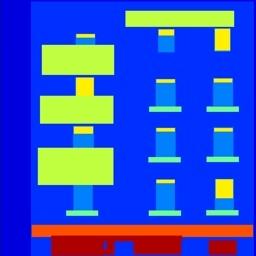} &
\includegraphics[width=0.16666666666667\linewidth]{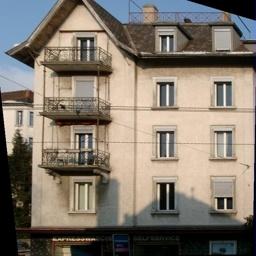} &
\includegraphics[width=0.16666666666667\linewidth]{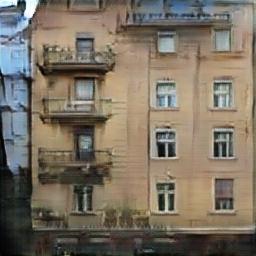}

\end{tabular} \egroup 
\end{center}
\caption{Example results of our method on facades labels$\rightarrow$photo, compared to ground truth.}
\label{facades_lotsofresults}
\end{figure*}

\begin{figure*}
\begin{center}
\bgroup 
 \def\arraystretch{0.2} 
 \setlength\tabcolsep{0.2pt}
\begin{tabular}{cccccc}
Input & Ground truth & Output & Input & Ground truth & Output \\ 
\includegraphics[width=0.16666666666667\linewidth]{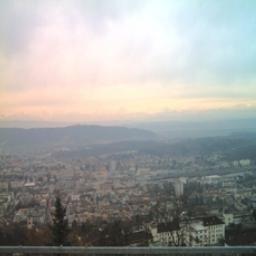} &
\includegraphics[width=0.16666666666667\linewidth]{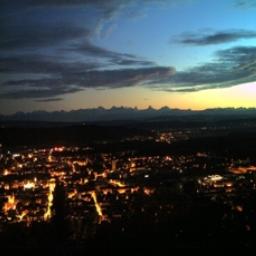} &
\includegraphics[width=0.16666666666667\linewidth]{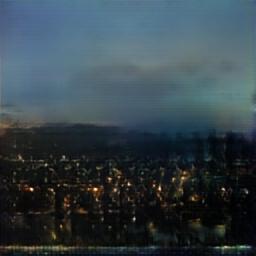} \hspace{0.025in} &
\includegraphics[width=0.16666666666667\linewidth]{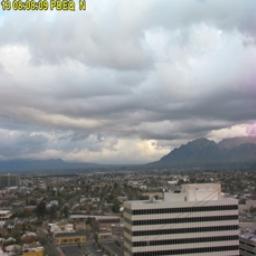} &
\includegraphics[width=0.16666666666667\linewidth]{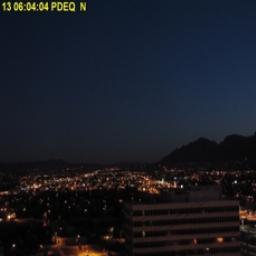} &
\includegraphics[width=0.16666666666667\linewidth]{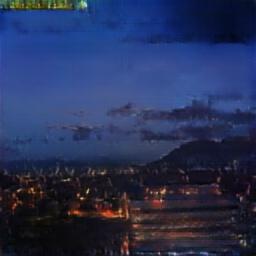} \\ 
\includegraphics[width=0.16666666666667\linewidth]{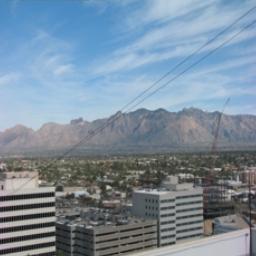} &
\includegraphics[width=0.16666666666667\linewidth]{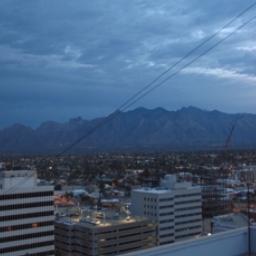} &
\includegraphics[width=0.16666666666667\linewidth]{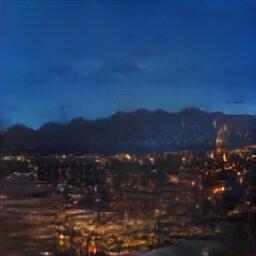} \hspace{0.025in} &
\includegraphics[width=0.16666666666667\linewidth]{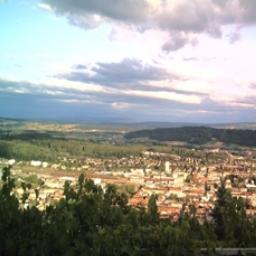} &
\includegraphics[width=0.16666666666667\linewidth]{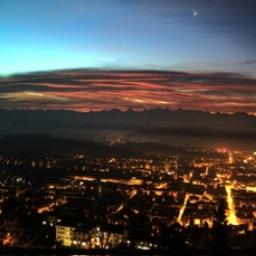} &
\includegraphics[width=0.16666666666667\linewidth]{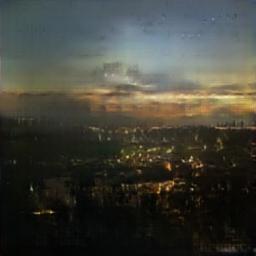} \\ 
\includegraphics[width=0.16666666666667\linewidth]{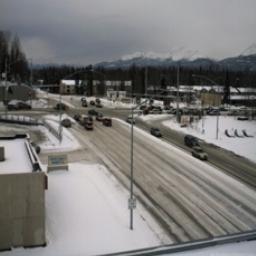} &
\includegraphics[width=0.16666666666667\linewidth]{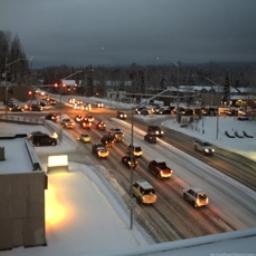} &
\includegraphics[width=0.16666666666667\linewidth]{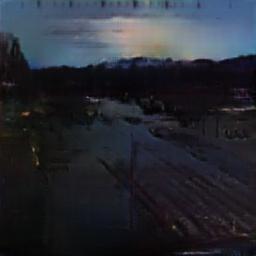} \hspace{0.025in} &
\includegraphics[width=0.16666666666667\linewidth]{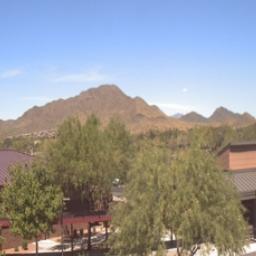} &
\includegraphics[width=0.16666666666667\linewidth]{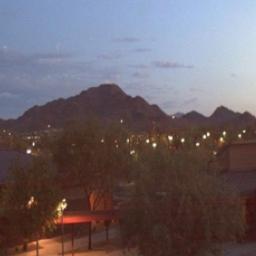} &
\includegraphics[width=0.16666666666667\linewidth]{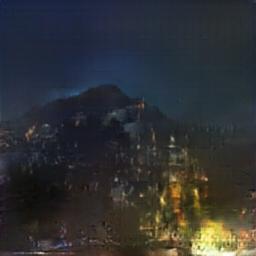} 

\end{tabular} \egroup 
\end{center}
\caption{Example results of our method on day$\rightarrow$night, compared to ground truth.}
\label{day2night_lotsofresults}
\end{figure*}

\begin{figure*}
\begin{center}
\bgroup 
 \def\arraystretch{0.2} 
 \setlength\tabcolsep{0.2pt}
\begin{tabular}{cccccc}
Input & Ground truth & Output & Input & Ground truth & Output \\ 
\includegraphics[width=0.16666666666667\linewidth]{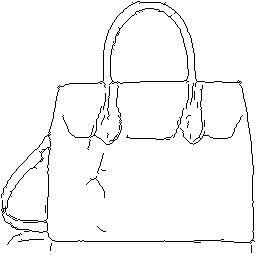} &
\includegraphics[width=0.16666666666667\linewidth]{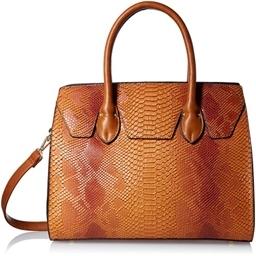} &
\includegraphics[width=0.16666666666667\linewidth]{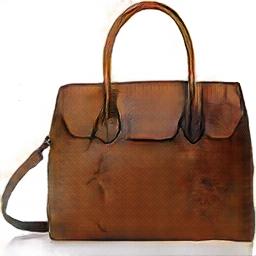} \hspace{0.025in} &
\includegraphics[width=0.16666666666667\linewidth]{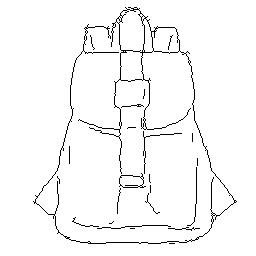} &
\includegraphics[width=0.16666666666667\linewidth]{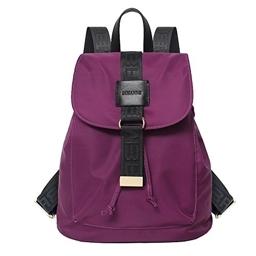} &
\includegraphics[width=0.16666666666667\linewidth]{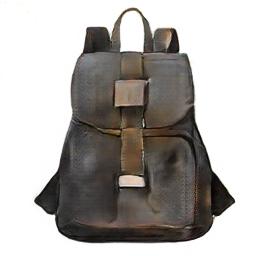} \\ 
\includegraphics[width=0.16666666666667\linewidth]{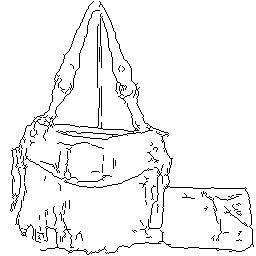} &
\includegraphics[width=0.16666666666667\linewidth]{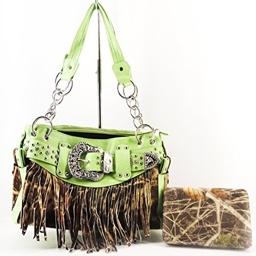} &
\includegraphics[width=0.16666666666667\linewidth]{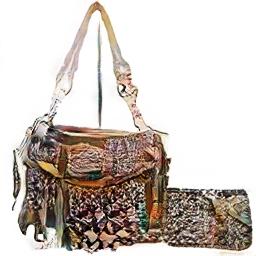} \hspace{0.025in} &
\includegraphics[width=0.16666666666667\linewidth]{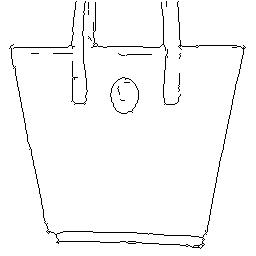} &
\includegraphics[width=0.16666666666667\linewidth]{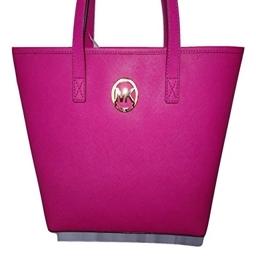} &
\includegraphics[width=0.16666666666667\linewidth]{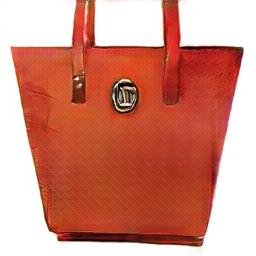} \\ 
\includegraphics[width=0.16666666666667\linewidth]{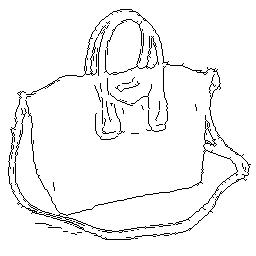} &
\includegraphics[width=0.16666666666667\linewidth]{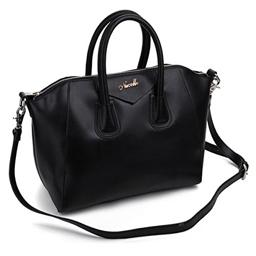} &
\includegraphics[width=0.16666666666667\linewidth]{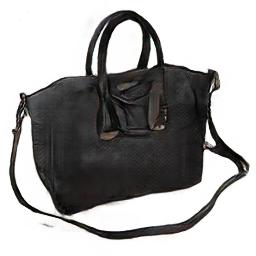} \hspace{0.025in} &
\includegraphics[width=0.16666666666667\linewidth]{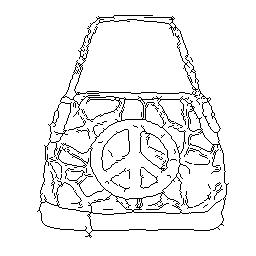} &
\includegraphics[width=0.16666666666667\linewidth]{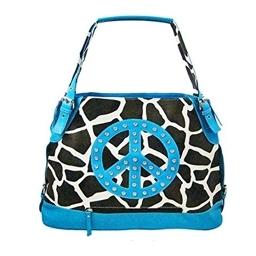} &
\includegraphics[width=0.16666666666667\linewidth]{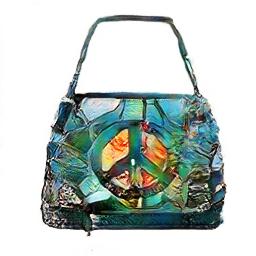}

\end{tabular} \egroup 
\end{center}
\caption{Example results of our method on automatically detected edges$\rightarrow$handbags, compared to ground truth.}
\label{handbags_edges_lotsofresults}
\end{figure*}

\begin{figure*}
\begin{center}
\bgroup 
 \def\arraystretch{0.2} 
 \setlength\tabcolsep{0.2pt}
\begin{tabular}{cccccc}
Input & Ground truth & Output & Input & Ground truth & Output \\ 
\includegraphics[width=0.16666666666667\linewidth]{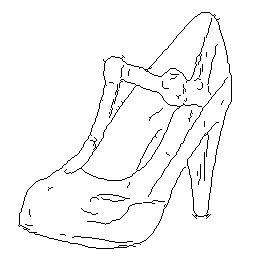} &
\includegraphics[width=0.16666666666667\linewidth]{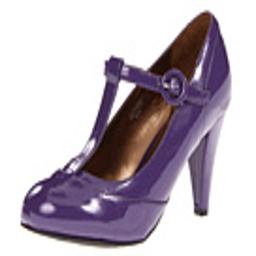} &
\includegraphics[width=0.16666666666667\linewidth]{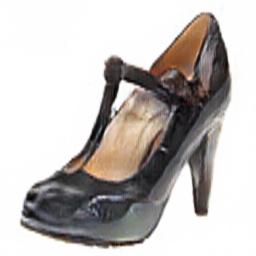} \hspace{0.025in} &
\includegraphics[width=0.16666666666667\linewidth]{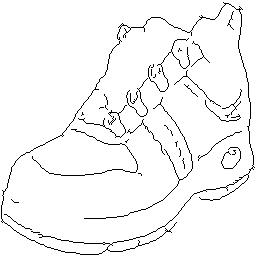} &
\includegraphics[width=0.16666666666667\linewidth]{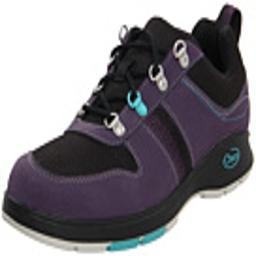} &
\includegraphics[width=0.16666666666667\linewidth]{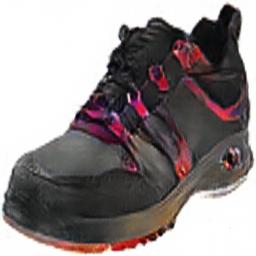} \\ 
\includegraphics[width=0.16666666666667\linewidth]{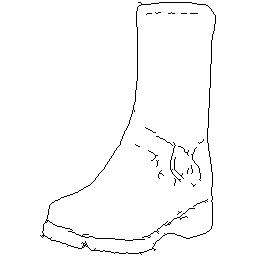} &
\includegraphics[width=0.16666666666667\linewidth]{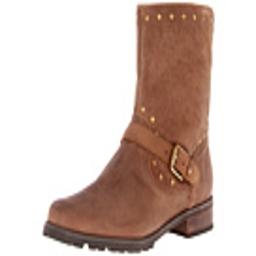} &
\includegraphics[width=0.16666666666667\linewidth]{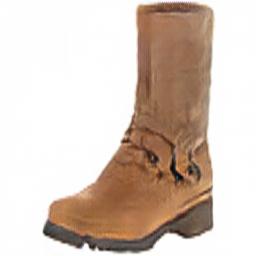} &
\includegraphics[width=0.16666666666667\linewidth]{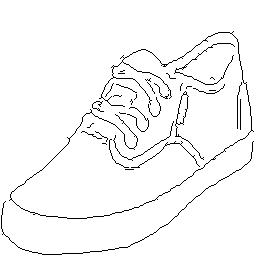} &
\includegraphics[width=0.16666666666667\linewidth]{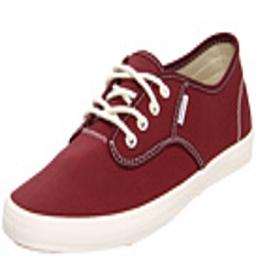} &
\includegraphics[width=0.16666666666667\linewidth]{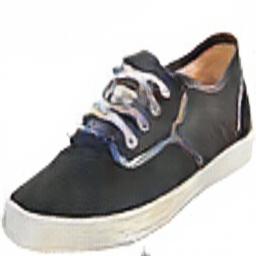} \\ 

\end{tabular} \egroup 
\end{center}
\caption{Example results of our method on automatically detected edges$\rightarrow$shoes, compared to ground truth.}
\label{shoes_edges_lotsofresults}
\end{figure*}

\begin{figure*}
\begin{center}
\bgroup 
 \def\arraystretch{0.2} 
 \setlength\tabcolsep{0.2pt}
\begin{tabular}{cccccc}
Input & Output & Input & Output & Input & Output \\ 

\includegraphics[width=0.167\linewidth]{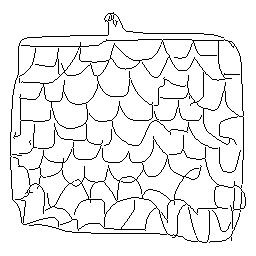} &
\includegraphics[width=0.167\linewidth]{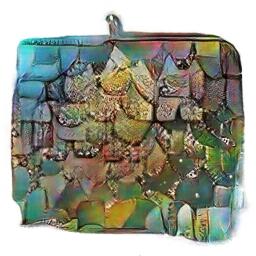} \hspace{0.025in} &
\includegraphics[width=0.167\linewidth]{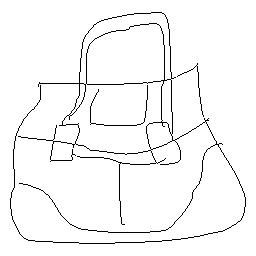} &
\includegraphics[width=0.167\linewidth]{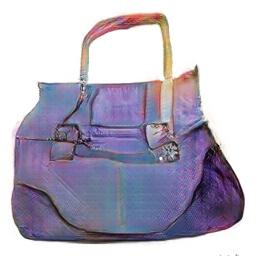} \hspace{0.025in} &

\includegraphics[width=0.167\linewidth]{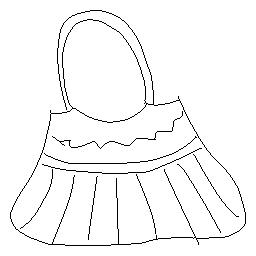} &
\includegraphics[width=0.167\linewidth]{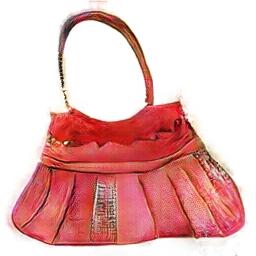} \\
\includegraphics[width=0.167\linewidth]{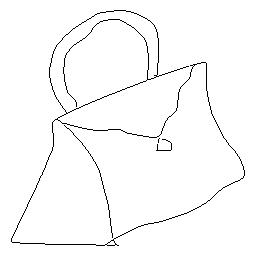} &
\includegraphics[width=0.167\linewidth]{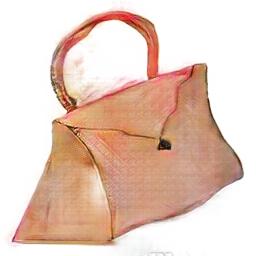} \hspace{0.025in} &

\includegraphics[width=0.167\linewidth]{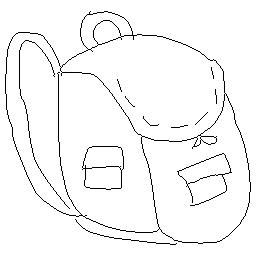} &
\includegraphics[width=0.167\linewidth]{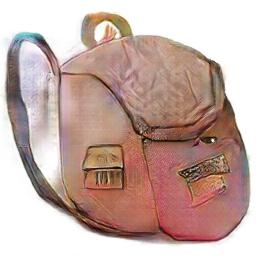} \hspace{0.025in} &
\includegraphics[width=0.167\linewidth]{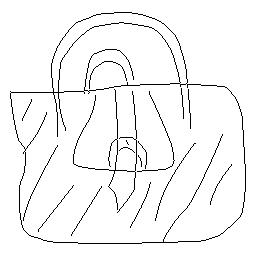} &
\includegraphics[width=0.167\linewidth]{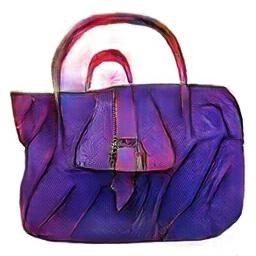} \\

\includegraphics[width=0.167\linewidth]{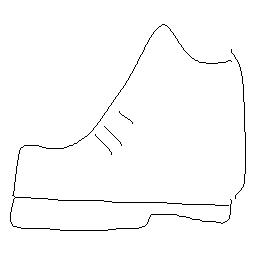} &
\includegraphics[width=0.167\linewidth]{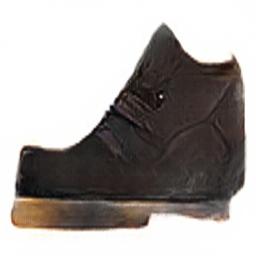} \hspace{0.025in} &
\includegraphics[width=0.167\linewidth]{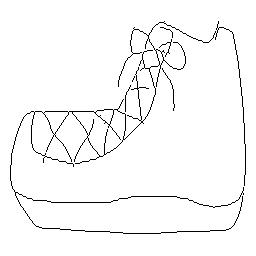} &
\includegraphics[width=0.167\linewidth]{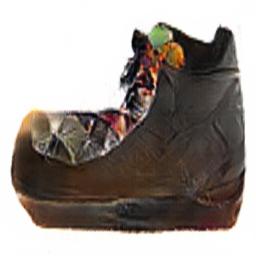} \hspace{0.025in} &

\includegraphics[width=0.167\linewidth]{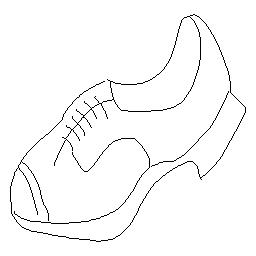} &
\includegraphics[width=0.167\linewidth]{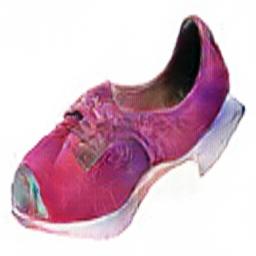} \\
\includegraphics[width=0.167\linewidth]{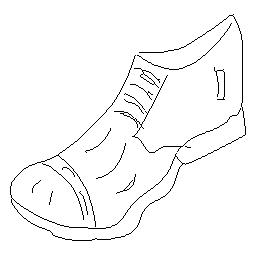} &
\includegraphics[width=0.167\linewidth]{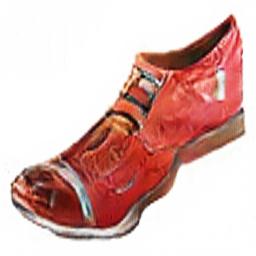} \hspace{0.025in} &

\includegraphics[width=0.167\linewidth]{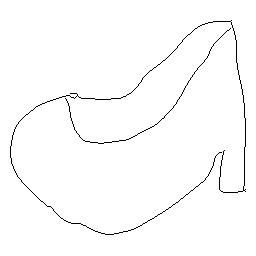} &
\includegraphics[width=0.167\linewidth]{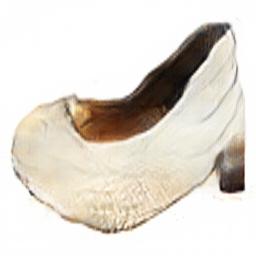} \hspace{0.025in} &
\includegraphics[width=0.167\linewidth]{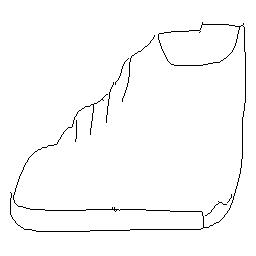} &
\includegraphics[width=0.167\linewidth]{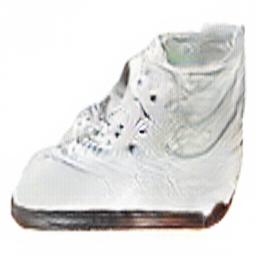}

\end{tabular} \egroup 
\end{center}
\caption{Additional results of the edges$\rightarrow$photo models applied to human-drawn sketches from \cite{eitz2012humans}. Note that the models were trained on automatically detected edges, but generalize to human drawings}
\label{sketches_lotsofresults}
\end{figure*}

\begin{figure*}
    \centering
    \includegraphics[width=0.49\hsize]{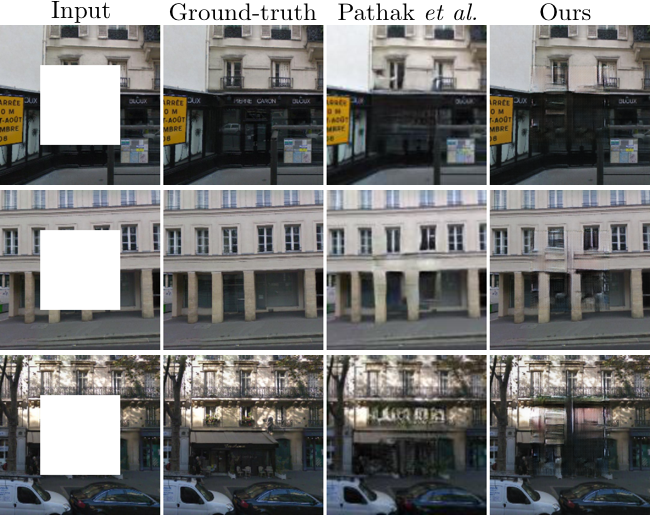}
    \includegraphics[width=0.49\hsize]{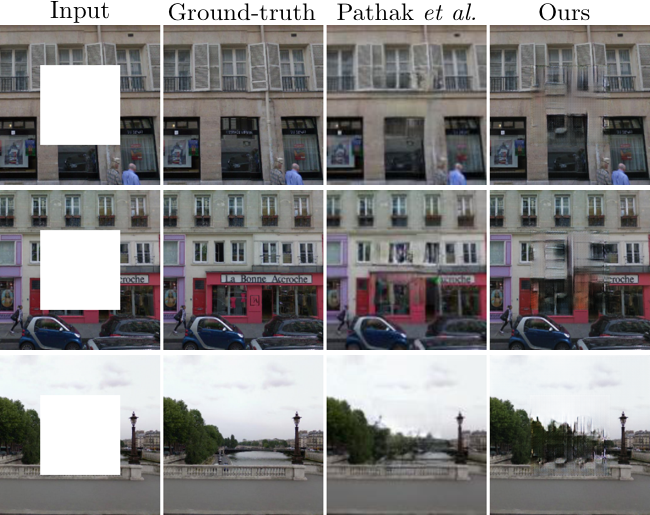}
    \caption{Example results on photo inpainting, compared to \cite{pathak2016context}, on the Paris StreetView dataset \cite{doersch2012makes}. This experiment demonstrates that the U-net architecture can be effective even when the predicted pixels are not geometrically aligned with the information in the input -- the information used to fill in the central hole has to be found in the periphery of these photos.}
    \label{inpaint}
\end{figure*}

\begin{figure*}
\begin{center}
\bgroup 
 \def\arraystretch{0.2} 
 \setlength\tabcolsep{0.2pt}
\begin{tabular}{cccccc}
Input & Ground truth & Output & Input & Ground truth & Output \\ 
\includegraphics[width=0.16666666666667\linewidth]{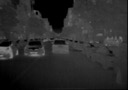} &
\includegraphics[width=0.16666666666667\linewidth]{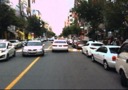} &
\includegraphics[width=0.16666666666667\linewidth]{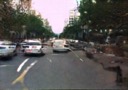} \hspace{0.01in} &
\includegraphics[width=0.16666666666667\linewidth]{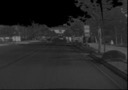} &
\includegraphics[width=0.16666666666667\linewidth]{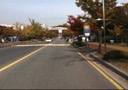} &
\includegraphics[width=0.16666666666667\linewidth]{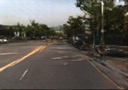} \\ 
\includegraphics[width=0.16666666666667\linewidth]{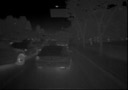} &
\includegraphics[width=0.16666666666667\linewidth]{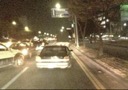} &
\includegraphics[width=0.16666666666667\linewidth]{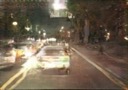} \hspace{0.01in} &
\includegraphics[width=0.16666666666667\linewidth]{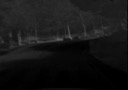} &
\includegraphics[width=0.16666666666667\linewidth]{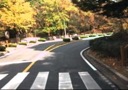} &
\includegraphics[width=0.16666666666667\linewidth]{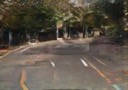} 
\end{tabular} \egroup 
\end{center}
\caption{Example results on translating thermal images to RGB photos, on the dataset from \cite{hwang2015multispectral}.}
\label{thermal2rgb}
\end{figure*}

\begin{figure*}
 \centering
 \includegraphics[width=1.0\hsize]{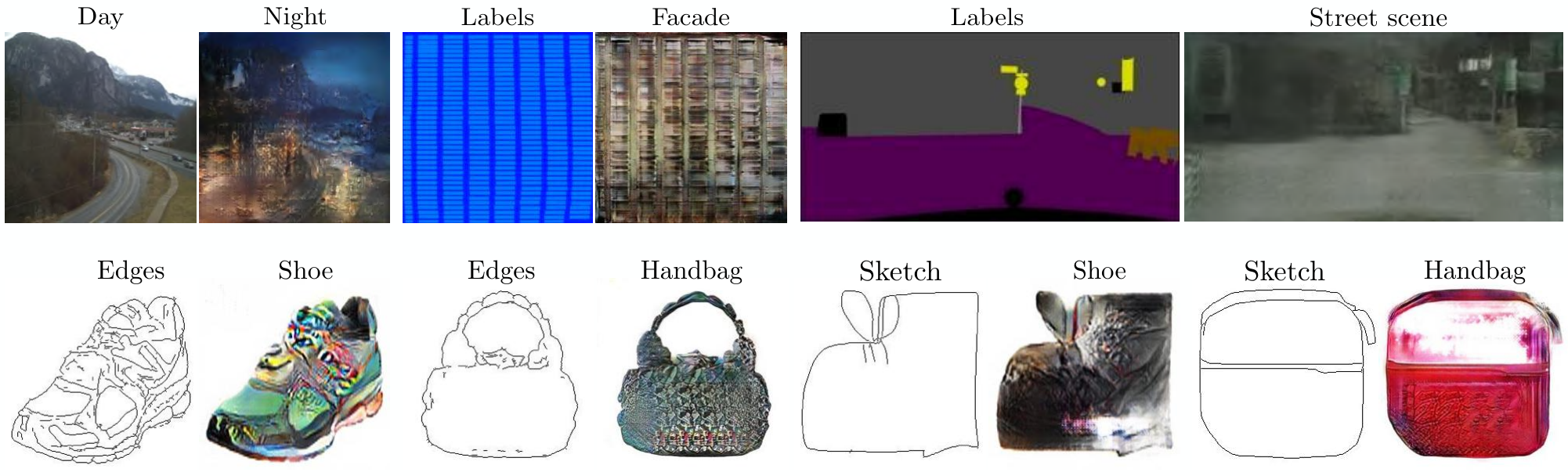}
 \vspace{-0.2in}
  \caption{Example failure cases. Each pair of images shows input on the left and output on the right. These examples are selected as some of the worst results on our tasks. Common failures include artifacts in regions where the input image is sparse, and difficulty in handling unusual inputs. Please see \texttt{https://phillipi.github.io/pix2pix/} for more comprehensive results.}
 \label{failure_cases}
\end{figure*}

Details of training on each of these datasets are provided in the supplemental materials online. In all cases, the input and output are simply 1-3 channel images. Qualitative results are shown in Figures \ref{sat2map_res}, \ref{colorization_res}, \ref{twitter}, \ref{cityscapes_image_to_labels}, \ref{cityscapes_lotsofresults}, \ref{facades_lotsofresults},
\ref{day2night_lotsofresults}, \ref{handbags_edges_lotsofresults}, \ref{shoes_edges_lotsofresults}, \ref{sketches_lotsofresults}, \ref{inpaint}, \ref{thermal2rgb}. Several failure cases are highlighted in Figure \ref{failure_cases}. More comprehensive results are available at \texttt{https://phillipi.github.io/pix2pix/}.

{\bf Data requirements and speed} We note that decent results can often be obtained even on small datasets. Our facade training set consists of just 400 images (see results in Figure \ref{facades_lotsofresults}), and the day to night training set consists of only 91 unique webcams (see results in Figure \ref{day2night_lotsofresults}). On datasets of this size, training can be very fast: for example, the results shown in Figure \ref{facades_lotsofresults} took less than two hours of training on a single Pascal Titan X GPU. At test time, all models run in well under a second on this GPU.

\begin{figure*}[t]
\begin{center}
\bgroup 
 \def\arraystretch{0.2} 
 \setlength\tabcolsep{0.2pt}
\begin{tabular}{ccccc}
Input & Ground truth & L1 & cGAN & L1 + cGAN \\ 
\includegraphics[width=0.2\linewidth]{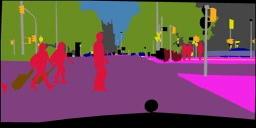} &
\includegraphics[width=0.2\linewidth]{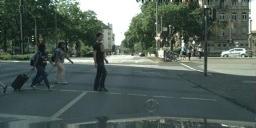} &
\includegraphics[width=0.2\linewidth]{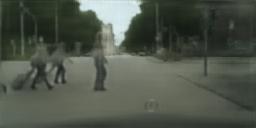} &
\includegraphics[width=0.2\linewidth]{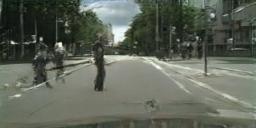} &
\includegraphics[width=0.2\linewidth]{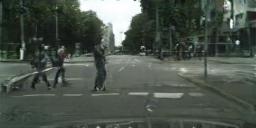} \\ 
\includegraphics[width=0.2\linewidth]{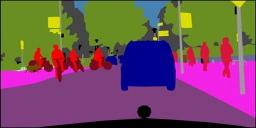} &
\includegraphics[width=0.2\linewidth]{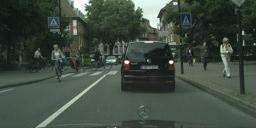} &
\includegraphics[width=0.2\linewidth]{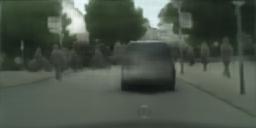} &
\includegraphics[width=0.2\linewidth]{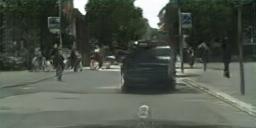} &
\includegraphics[width=0.2\linewidth]{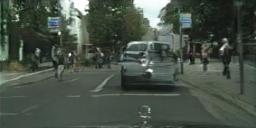} \\

\includegraphics[width=0.2\linewidth]{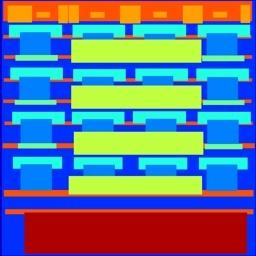} &
\includegraphics[width=0.2\linewidth]{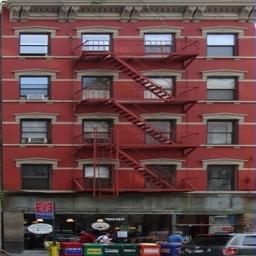} &
\includegraphics[width=0.2\linewidth]{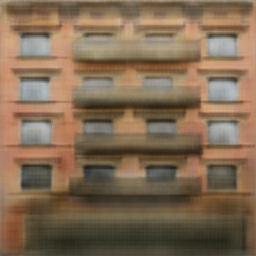} &
\includegraphics[width=0.2\linewidth]{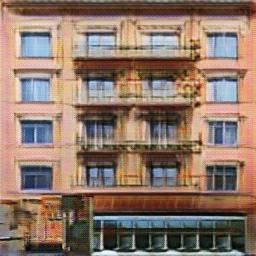} &
\includegraphics[width=0.2\linewidth]{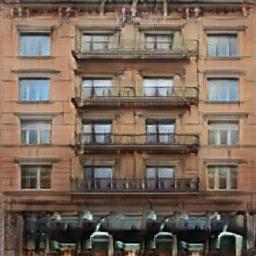} \\ 

\includegraphics[width=0.2\linewidth]{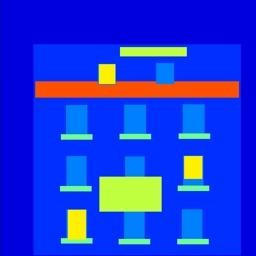} &
\includegraphics[width=0.2\linewidth]{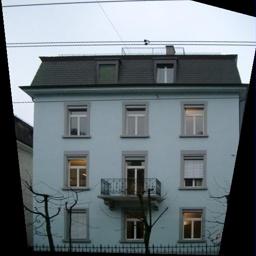} &
\includegraphics[width=0.2\linewidth]{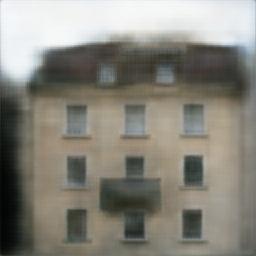} &
\includegraphics[width=0.2\linewidth]{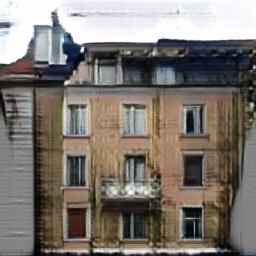} &
\includegraphics[width=0.2\linewidth]{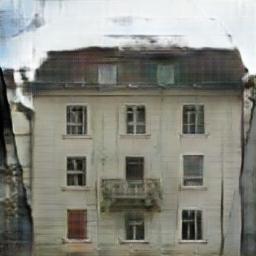}

\end{tabular} \egroup 
\end{center}
\vspace{-0.2in}
\caption{Different losses induce different quality of results. Each column shows results trained under a different loss. Please see \texttt{https://phillipi.github.io/pix2pix/} for additional examples.}
\label{cityscapes_loss_variations_qualitative}
\vspace{-0.2in}
\end{figure*}

\subsection{Evaluation metrics}
Evaluating the quality of synthesized images is an open and difficult problem \cite{salimans2016improved}. Traditional metrics such as per-pixel mean-squared error do not assess joint statistics of the result, and therefore do not measure the very structure that structured losses aim to capture.

To more holistically evaluate the visual quality of our results, we employ two tactics. First, we run ``real vs. fake" perceptual studies on Amazon Mechanical Turk (AMT). For graphics problems like colorization and photo generation, plausibility to a human observer is often the ultimate goal. Therefore, we test our map generation, aerial photo generation, and image colorization using this approach.

Second, we measure whether or not our synthesized cityscapes are realistic enough that off-the-shelf recognition system can recognize the objects in them. This metric is similar to the ``inception score" from \cite{salimans2016improved}, the object detection evaluation in \cite{wang2016generative}, and the ``semantic interpretability" measures in \cite{zhang2016colorful} and \cite{owens2016visually}.

{\bf AMT perceptual studies} For our AMT experiments, we followed the protocol from \cite{zhang2016colorful}: Turkers were presented with a series of trials that pitted a ``real" image against a ``fake" image generated by our algorithm. On each trial, each image appeared for 1 second, after which the images disappeared and Turkers were given unlimited time to respond as to which was fake. The first 10 images of each session were practice and Turkers were given feedback. No feedback was provided on the 40 trials of the main experiment. Each session tested just one algorithm at a time, and Turkers were not allowed to complete more than one session. $\sim 50$ Turkers evaluated each algorithm. Unlike \cite{zhang2016colorful}, we did not include vigilance trials. For our colorization experiments, the real and fake images were generated from the same grayscale input. For map$\leftrightarrow$aerial photo, the real and fake images were not generated from the same input, in order to make the task more difficult and avoid floor-level results. For map$\leftrightarrow$aerial photo, we trained on $256 \times 256$ resolution images, but exploited fully-convolutional translation (described above) to test on $512 \times 512$ images, which were then downsampled and presented to Turkers at $256 \times 256$ resolution. For colorization, we trained and tested on $256 \times 256$ resolution images and presented the results to Turkers at this same resolution.

{\bf ``FCN-score"} While quantitative evaluation of generative models is known to be challenging, recent works~\cite{salimans2016improved, wang2016generative, zhang2016colorful,owens2016visually} have tried using pre-trained semantic classifiers to measure the discriminability of the generated stimuli as a pseudo-metric. The intuition is that if the generated images are realistic, classifiers trained on real images will be able to classify the synthesized image correctly as well. To this end, we adopt the popular FCN-8s~\cite{long2015fully} architecture for semantic segmentation, and train it on the cityscapes dataset. We then score synthesized photos by the classification accuracy against the labels these photos were synthesized from.

\begin{figure}
 \centering
 \includegraphics[width=1.0\hsize]{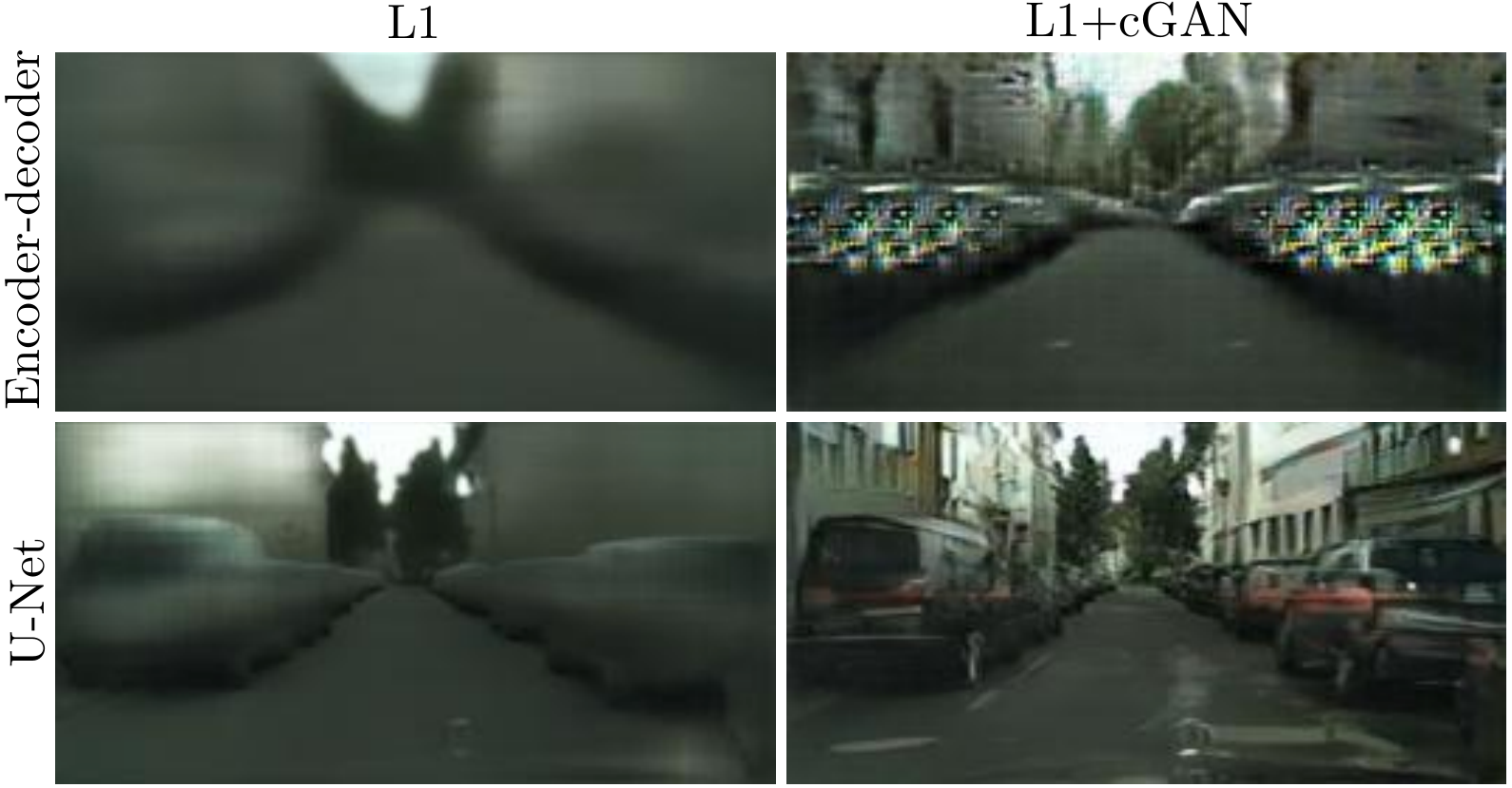}
 \vspace{-0.2in}
  \caption{Adding skip connections to an encoder-decoder to create a ``U-Net" results in much higher quality results.}
 \label{enc_dec_vs_unet}
 \vspace{-0.2in}
\end{figure}

\begin{figure*}
\begin{center}
\bgroup 
 \def\arraystretch{0.2} 
 \setlength\tabcolsep{0.2pt}
\begin{tabular}{ccccc}
L1 & 1$\times$1 & 16$\times$16 & 70$\times$70 & 286$\times$286 \\ 
\includegraphics[width=0.2\linewidth]{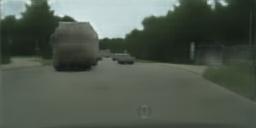} &
\includegraphics[width=0.2\linewidth]{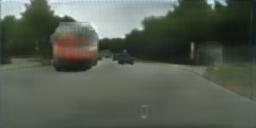} &
\includegraphics[width=0.2\linewidth]{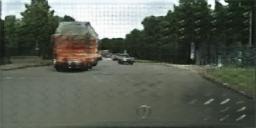} &
\includegraphics[width=0.2\linewidth]{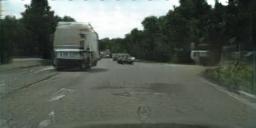} &
\includegraphics[width=0.2\linewidth]{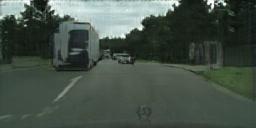} 
\end{tabular} \egroup 
\end{center}
\vspace{-0.1in}
\caption{Patch size variations. Uncertainty in the output manifests itself differently for different loss functions. Uncertain regions become blurry and desaturated under L1. The 1x1 PixelGAN encourages greater color diversity but has no effect on spatial statistics. The 16x16 PatchGAN creates locally sharp results, but also leads to tiling artifacts beyond the scale it can observe. The 70$\times$70 PatchGAN forces outputs that are sharp, even if incorrect, in both the spatial and spectral (colorfulness) dimensions. The full 286$\times$286 ImageGAN produces results that are visually similar to the 70$\times$70 PatchGAN, but somewhat lower quality according to our FCN-score metric (Table \ref{tab:patchsize_variations}). Please see \texttt{https://phillipi.github.io/pix2pix/} for additional examples.}
\label{patchsize_variations_qualitative}
\vspace{-0.2in}
\end{figure*}

\begin{table}
\centering
\scalebox{0.75} {
\begin{tabular}{lcccc}
 & & & \\
\textbf{Loss} & \textbf{Per-pixel acc.} & \textbf{Per-class acc.} & \textbf{Class IOU} \\ \hline
\textbf{L1} & 0.42 & 0.15 & 0.11 \\
\textbf{GAN} & 0.22 & 0.05 & 0.01 \\
\textbf{cGAN} & 0.57 & 0.22 & 0.16 \\
\textbf{L1+GAN} & 0.64 & 0.20 & 0.15 \\
\textbf{L1+cGAN} & \textbf{0.66} & \textbf{0.23} & \textbf{0.17} \\ \hline
\textbf{Ground truth} & 0.80 & 0.26 & 0.21 \\
\end{tabular} }
\vspace{-0.1in}
\caption {FCN-scores for different losses, evaluated on Cityscapes labels$\leftrightarrow$photos.}
\label{tab:loss_variations}
\vspace{-0.1in}
\end{table}

\begin{table}
\centering
\scalebox{0.75} {
\begin{tabular}{lcccc}
 & & & \\
\textbf{Loss} & \textbf{Per-pixel acc.} & \textbf{Per-class acc.} & \textbf{Class IOU} \\ \hline
\textbf{Encoder-decoder (L1)} & 0.35 & 0.12 & 0.08 \\
\textbf{Encoder-decoder (L1+cGAN)} & 0.29 & 0.09 & 0.05 \\
\textbf{U-net (L1)} & 0.48 & 0.18 & 0.13 \\
\textbf{U-net (L1+cGAN)} & \textbf{0.55} & \textbf{0.20} & \textbf{0.14} \\
\end{tabular} }
\vspace{-0.1in}
\caption {FCN-scores for different generator architectures (and objectives), evaluated on Cityscapes labels$\leftrightarrow$photos. (U-net (L1-cGAN) scores differ from those reported in other tables since batch size was 10 for this experiment and 1 for other tables, and random variation between training runs.)}
\label{tab:arch_variations}
\vspace{-0.1in}
\end{table}

\begin{table}
\centering
\scalebox{0.75} {
\begin{tabular}{lccccc}
\textbf{Discriminator} & & & \\
\textbf{receptive field} & \textbf{Per-pixel acc.} & \textbf{Per-class acc.} & \textbf{Class IOU} \\ \hline
\textbf{1$\times$1} & 0.39 & 0.15 & 0.10 \\
\textbf{16$\times$16} & 0.65 & 0.21 & \textbf{0.17}  \\
\textbf{70$\times$70} & \textbf{0.66} & \textbf{0.23} & \textbf{0.17}  \\
\textbf{286$\times$286} & 0.42 & 0.16 & 0.11 \\
\end{tabular} }
\vspace{-0.1in}
\caption {FCN-scores for different receptive field sizes of the discriminator, evaluated on Cityscapes labels$\rightarrow$photos. Note that input images are $256 \times 256$ pixels and larger receptive fields are padded with zeros.}
\vspace{-0.2in}
\label{tab:patchsize_variations}
\end{table}

\begin{figure*}[t]
\begin{center}
\bgroup 
\begin{tabular}{cccc}
\subfloat[][]{\includegraphics[width=0.2\linewidth]{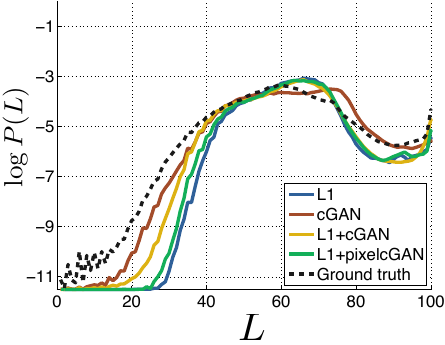}} &
\subfloat[][]{\includegraphics[width=0.2\linewidth]{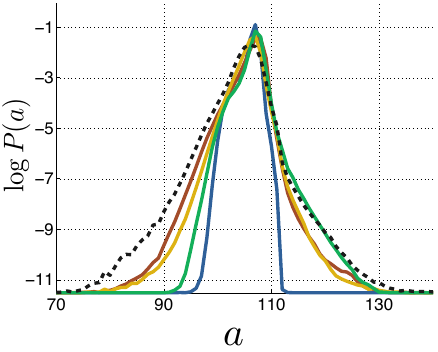}} &
\subfloat[][]{\includegraphics[width=0.2\linewidth]{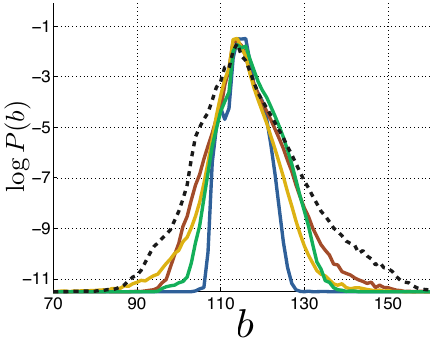}} &

\subfloat[][]{
\centering
\scalebox{0.75} {
\begin{tabular}[b]{lccccc}
& \multicolumn{3}{c}{Histogram intersection} \\
& \multicolumn{3}{c}{against ground truth} \\
\textbf{Loss} & L & a & b \\ \hline
\textbf{L1} & 0.81 & 0.69 & 0.70 \\
\textbf{cGAN} & \textbf{0.87} & 0.74 & \textbf{0.84} \\
\textbf{L1+cGAN} & 0.86 & \textbf{0.84} & 0.82 \\
\hline
\textbf{PixelGAN} & 0.83 & 0.68 & 0.78 
\end{tabular} }
\label{tab:color_hists}
}

\end{tabular} \egroup 
\end{center}
\vspace{-0.2in}
\caption{Color distribution matching property of the cGAN, tested on Cityscapes. (c.f. Figure 1 of the original GAN paper \cite{goodfellow2014generative}). Note that the histogram intersection scores are dominated by differences in the high probability region, which are imperceptible in the plots, which show log probability and therefore emphasize differences in the low probability regions.}
\label{color_hists}
\end{figure*}

\subsection{Analysis of the objective function}

Which components of the objective in Eqn. \ref{full_objective} are important? We run ablation studies to isolate the effect of the L1 term, the GAN term, and to compare using a discriminator conditioned on the input (cGAN, Eqn. \ref{cGAN_equation}) against using an unconditional discriminator (GAN, Eqn. \ref{GAN_equation}).

Figure \ref{cityscapes_loss_variations_qualitative} shows the qualitative effects of these variations on two labels$\rightarrow$photo problems. L1 alone leads to reasonable but blurry results. The cGAN alone (setting $\lambda=0$ in Eqn. \ref{full_objective}) gives much sharper results but introduces visual artifacts on certain applications. Adding both terms together (with $\lambda=100$) reduces these artifacts.

We quantify these observations using the FCN-score on the cityscapes labels$\rightarrow$photo task (Table \ref{tab:loss_variations}): the GAN-based objectives achieve higher scores, indicating that the synthesized images include more recognizable structure. We also test the effect of removing conditioning from the discriminator (labeled as GAN). In this case, the loss does not penalize mismatch between the input and output; it only cares that the output look realistic. This variant results in poor performance; examining the results reveals that the generator collapsed into producing nearly the exact same output regardless of input photograph. Clearly, it is important, in this case, that the loss measure the quality of the match between input and output, and indeed cGAN performs much better than GAN. Note, however, that adding an L1 term also encourages that the output respect the input, since the L1 loss penalizes the distance between ground truth outputs, which correctly match the input, and synthesized outputs, which may not. Correspondingly, L1+GAN is also effective at creating realistic renderings that respect the input label maps. Combining all terms, L1+cGAN, performs similarly well.

{\bf Colorfulness} A striking effect of conditional GANs is that they produce sharp images, hallucinating spatial structure even where it does not exist in the input label map. One might imagine cGANs have a similar effect on ``sharpening" in the spectral dimension -- i.e. making images more colorful. Just as L1 will incentivize a blur when it is uncertain where exactly to locate an edge, it will also incentivize an average, grayish color when it is uncertain which of several plausible color values a pixel should take on. Specially, L1 will be minimized by choosing the median of the conditional probability density function over possible colors. An adversarial loss, on the other hand, can in principle become aware that grayish outputs are unrealistic, and encourage matching the true color distribution \cite{goodfellow2014generative}. In Figure \ref{color_hists}, we investigate whether our cGANs actually achieve this effect on the Cityscapes dataset. The plots show the marginal distributions over output color values in Lab color space. The ground truth distributions are shown with a dotted line. It is apparent that L1 leads to a narrower distribution than the ground truth, confirming the hypothesis that L1 encourages average, grayish colors. Using a cGAN, on the other hand, pushes the output distribution closer to the ground truth.

\subsection{Analysis of the generator architecture}\label{analysis_of_gen_arch}

A U-Net architecture allows low-level information to shortcut across the network. Does this lead to better results? Figure \ref{enc_dec_vs_unet} and Table \ref{tab:arch_variations} compare the U-Net against an encoder-decoder on cityscape generation. The encoder-decoder is created simply by severing the skip connections in the U-Net. The encoder-decoder is unable to learn to generate realistic images in our experiments. The advantages of the U-Net appear not to be specific to conditional GANs: when both U-Net and encoder-decoder are trained with an L1 loss, the U-Net again achieves the superior results.

\subsection{From PixelGANs to PatchGANs to ImageGANs}\label{patch_variations_section}

We test the effect of varying the patch size $N$ of our discriminator receptive fields, from a $1\times1$ ``PixelGAN" to a full $286\times286$ ``ImageGAN"\footnote{We achieve this variation in patch size by adjusting the depth of the GAN discriminator. Details of this process, and the discriminator architectures, are provided in the in the supplemental materials online.}. Figure \ref{patchsize_variations_qualitative} shows qualitative results of this analysis and Table \ref{tab:patchsize_variations} quantifies the effects using the FCN-score. Note that elsewhere in this paper, unless specified, all experiments use $70\times70$ PatchGANs, and for this section all experiments use an L1+cGAN loss.

The PixelGAN has no effect on spatial sharpness but does increase the colorfulness of the results (quantified in Figure \ref{color_hists}). For example, the bus in Figure \ref{patchsize_variations_qualitative} is painted gray when the net is trained with an L1 loss, but becomes red with the PixelGAN loss. Color histogram matching is a common problem in image processing \cite{reinhard_color_2001}, and PixelGANs may be a promising lightweight solution.

Using a $16\times16$ PatchGAN is sufficient to promote sharp outputs, and achieves good FCN-scores, but also leads to tiling artifacts. The $70\times70$ PatchGAN alleviates these artifacts and achieves slightly better scores. Scaling beyond this, to the full $286\times286$ ImageGAN, does not appear to improve the visual quality of the results, and in fact gets a considerably lower FCN-score (Table \ref{tab:patchsize_variations}). This may be because the ImageGAN has many more parameters and greater depth than the $70\times70$ PatchGAN, and may be harder to train.

\begin{figure*}[t]
 \centering
 \includegraphics[width=1.0\hsize]{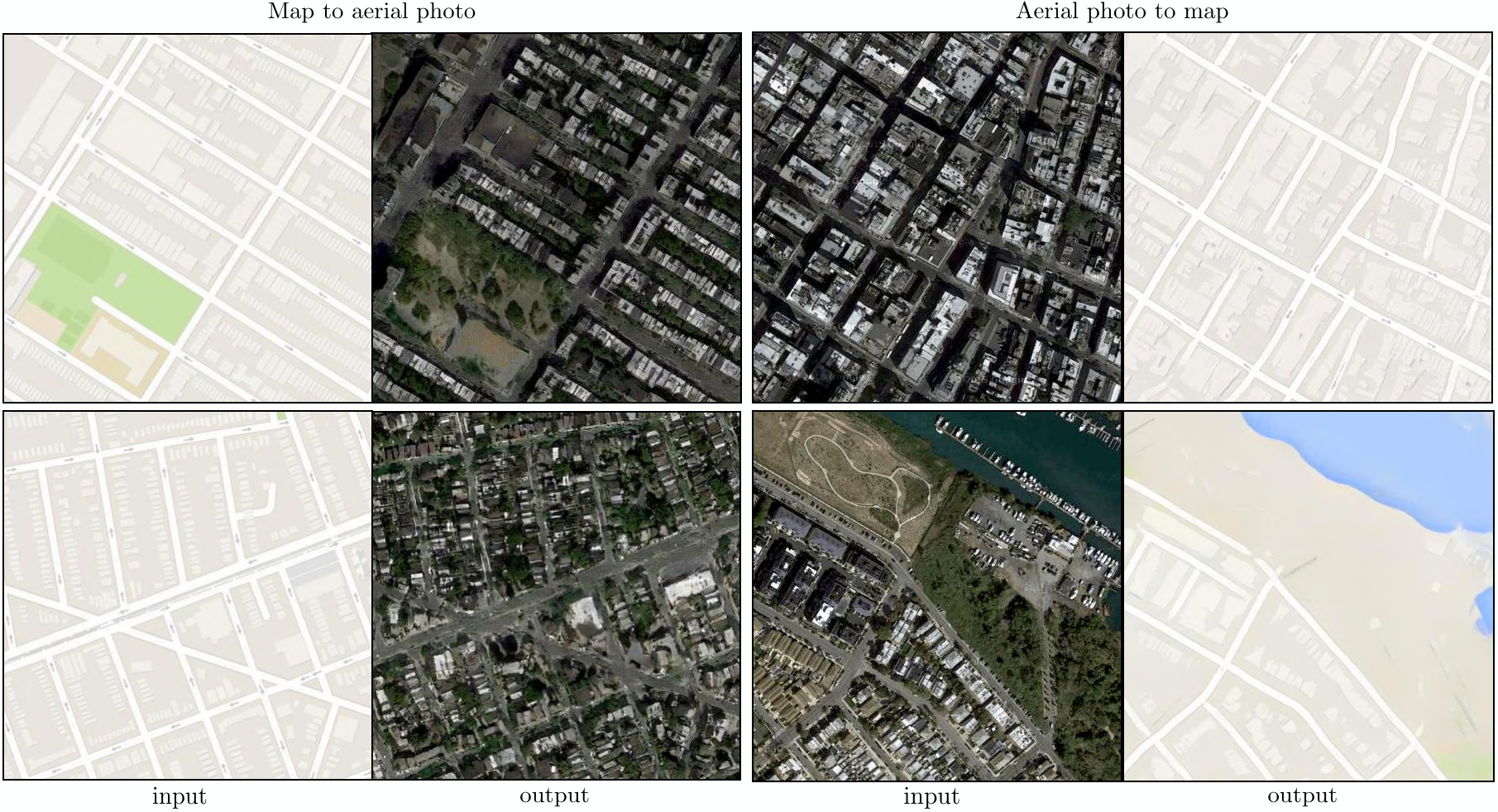}
 \vspace{-0.2in}
  \caption{Example results on Google Maps at 512x512 resolution (model was trained on images at $256 \times 256$ resolution, and run convolutionally on the larger images at test time). Contrast adjusted for clarity.}
 \label{sat2map_res}
 \vspace{-0.2in}
\end{figure*}

{\bf Fully-convolutional translation} An advantage of the PatchGAN is that a fixed-size patch discriminator can be applied to arbitrarily large images. We may also apply the generator convolutionally, on larger images than those on which it was trained. We test this on the map$\leftrightarrow$aerial photo task. After training a generator on $256\times256$ images, we test it on $512\times512$ images. The results in Figure \ref{sat2map_res} demonstrate the effectiveness of this approach.

\begin{figure}[h]

\bgroup
 \def\arraystretch{0.2}
 \setlength\tabcolsep{0.2pt}

\begin{tabular}{cccc}
Input & Ground truth & L1 & cGAN \\
\includegraphics[width=0.25\linewidth]{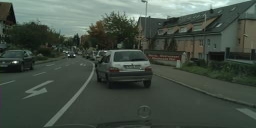} &
\includegraphics[width=0.25\linewidth]{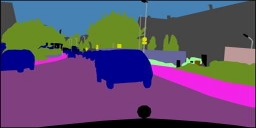} &
\includegraphics[width=0.25\linewidth]{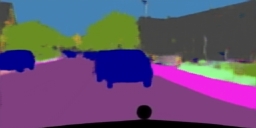} &
\includegraphics[width=0.25\linewidth]{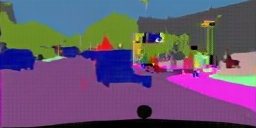} \\
\includegraphics[width=0.25\linewidth]{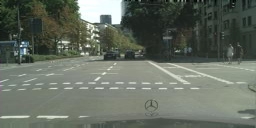} &
\includegraphics[width=0.25\linewidth]{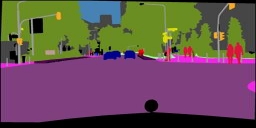} &
\includegraphics[width=0.25\linewidth]{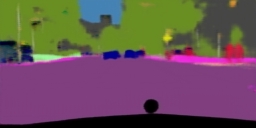} &
\includegraphics[width=0.25\linewidth]{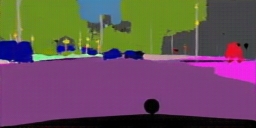} \\
\includegraphics[width=0.25\linewidth]{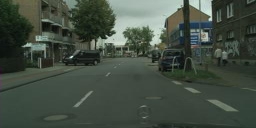} &
\includegraphics[width=0.25\linewidth]{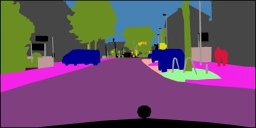} &
\includegraphics[width=0.25\linewidth]{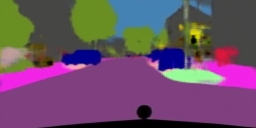} &
\includegraphics[width=0.25\linewidth]{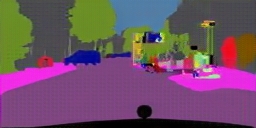} \\
\includegraphics[width=0.25\linewidth]{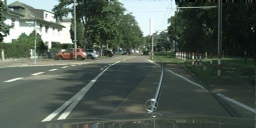} &
\includegraphics[width=0.25\linewidth]{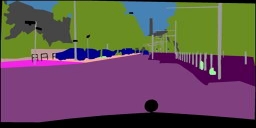} &
\includegraphics[width=0.25\linewidth]{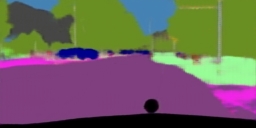} &
\includegraphics[width=0.25\linewidth]{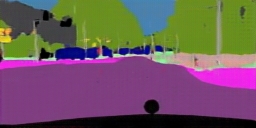}
\end{tabular} \egroup
\vspace{-0.1in}
\caption{Applying a conditional GAN to semantic segmentation. The cGAN produces sharp images that look at glance like the ground truth, but in fact include many small, hallucinated objects.}
\vspace{-0.2in}
\label{cityscapes_image_to_labels}
\end{figure}

\begin{table}
\centering
\scalebox{0.75} {
\begin{tabular}{lcc}
 & \textbf{Photo $\rightarrow$ Map}  &  \textbf{Map $\rightarrow$ Photo} \\
\textbf{Loss} & \% Turkers labeled \emph{real} & \% Turkers labeled \emph{real} \\
\hline
\textbf{L1} & 2.8\% $\pm$ 1.0\% & 0.8\% $\pm$ 0.3\% \\
\textbf{L1+cGAN} & 6.1\% $\pm$ 1.3\% & {\bf 18.9\% $\pm$ 2.5\%} \\
\end{tabular} }
\vspace{-0.1in}
\caption {AMT ``real vs fake" test on maps$\leftrightarrow$aerial photos.}
\vspace{-0.1in}
\label{tab:AMT_map2sat}
\end{table}

\begin{table}
\centering
\scalebox{0.75} {
\begin{tabular}{lc}
\textbf{Method} & \% Turkers labeled \emph{real} \\
\hline
\textbf{L2 regression from \cite{zhang2016colorful}} & 16.3\% $\pm$ 2.4\% \\
\textbf{Zhang et al. 2016 \cite{zhang2016colorful}} & \textbf{27.8\% $\pm$ 2.7\%} \\
\textbf{Ours} & 22.5\% $\pm$  1.6\% \\
\end{tabular} }
\vspace{-0.1in}
\caption {AMT ``real vs fake" test on colorization.}
\vspace{-0.1in}
\label{tab:AMT_colorization}
\end{table}

\subsection{Perceptual validation}

We validate the perceptual realism of our results on the tasks of map$\leftrightarrow$aerial photograph and grayscale$\rightarrow$color. Results of our AMT experiment for map$\leftrightarrow$photo are given in Table \ref{tab:AMT_map2sat}. The aerial photos generated by our method fooled participants on $18.9\%$ of trials, significantly above the L1 baseline, which produces blurry results and nearly never fooled participants. In contrast, in the photo$\rightarrow$map direction our method only fooled participants on $6.1$\% of trials, and this was not significantly different than the performance of the L1 baseline (based on bootstrap test). This may be because minor structural errors are more visible in maps, which have rigid geometry, than in aerial photographs, which are more chaotic.

\begin{figure*}[h]
    \centering
    \includegraphics[width=1.0\hsize]{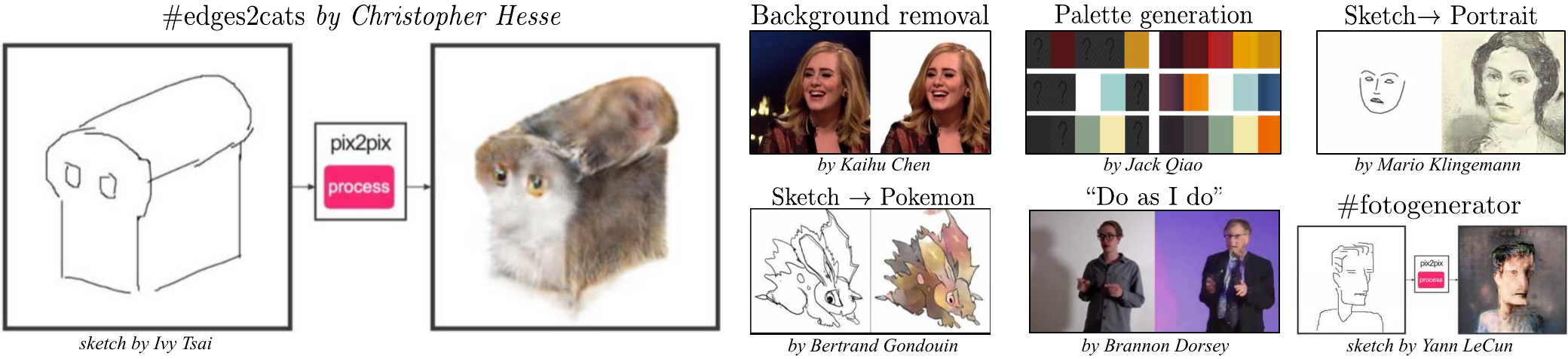}
    \vspace{-0.3in}
    \caption{Example applications developed by online community based on our {\tt pix2pix} codebase: \emph{\#edges2cats}~\cite{edges2cats} by Christopher Hesse, \emph{Background removal}~\cite{background} by Kaihu Chen, \emph{Palette generation}~\cite{palette} by Jack Qiao, \emph{Sketch $\rightarrow$ Portrait}~\cite{sketch2portrait} by Mario Klingemann, \emph{Sketch$\rightarrow$ Pokemon}~\cite{sketch2pokemon} by Bertrand Gondouin, \emph{``Do As I Do'' pose transfer}~\cite{pose} by Brannon Dorsey, and \emph{\#fotogenerator} by Bosman et al.~\cite{fotogenerator}.}
    \label{twitter}
    \vspace{-0.2in}
\end{figure*}

We trained colorization on ImageNet \cite{russakovsky2015imagenet}, and tested on the test split introduced by \cite{zhang2016colorful, larsson2016learning}. Our method, with L1+cGAN loss, fooled participants on $22.5\%$ of trials (Table \ref{tab:AMT_colorization}). We also tested the results of \cite{zhang2016colorful} and a variant of their method that used an L2 loss (see \cite{zhang2016colorful} for details). The conditional GAN scored similarly to the L2 variant of \cite{zhang2016colorful} (difference insignificant by bootstrap test), but fell short of \cite{zhang2016colorful}'s full method, which fooled participants on $27.8\%$ of trials in our experiment. We note that their method was specifically engineered to do well on colorization.

\subsection{Semantic segmentation}
Conditional GANs appear to be effective on problems where the output is highly detailed or photographic, as is common in image processing and graphics tasks. What about vision problems, like semantic segmentation, where the output is instead less complex than the input?

To begin to test this, we train a cGAN (with/without L1 loss) on cityscape photo$\rightarrow$labels. Figure \ref{cityscapes_image_to_labels} shows qualitative results, and quantitative classification accuracies are reported in Table \ref{tab:image_to_labels_results}. Interestingly, cGANs, trained \emph{without} the L1 loss, are able to solve this problem at a reasonable degree of accuracy. To our knowledge, this is the first demonstration of GANs successfully generating ``labels", which are nearly discrete, rather than ``images", with their continuous-valued variation\footnote{Note that the label maps we train on are not exactly discrete valued, as they are resized from the original maps using bilinear interpolation and saved as jpeg images, with some compression artifacts.}.
Although cGANs achieve some success, they are far from the best available method for solving this problem: simply using L1 regression gets better scores than using a cGAN, as shown in Table \ref{tab:image_to_labels_results}. We argue that for vision problems, the goal (i.e. predicting output close to the ground truth) may be less ambiguous than graphics tasks, and reconstruction losses like L1 are mostly sufficient.

\begin{table}
\centering
\scalebox{0.75} {
\begin{tabular}{lccccc}
 & & & \\
\textbf{Loss} & \textbf{Per-pixel acc.} & \textbf{Per-class acc.} & \textbf{Class IOU} \\ \hline
\textbf{L1} & \textbf{0.86} & \textbf{0.42} & \textbf{0.35} \\
\textbf{cGAN} & 0.74 & 0.28 & 0.22 \\
\textbf{L1+cGAN} & 0.83 & 0.36 & 0.29 \\ 
\end{tabular} }
\vspace{-0.1in}
\caption {Performance of photo$\rightarrow$labels on cityscapes.}
\vspace{-0.2in}
\label{tab:image_to_labels_results}
\end{table}

\subsection{Community-driven Research}

\begin{figure}[h]
\centering
    \href{https://vimeo.com/260612034}{\includegraphics[width=0.86\linewidth]{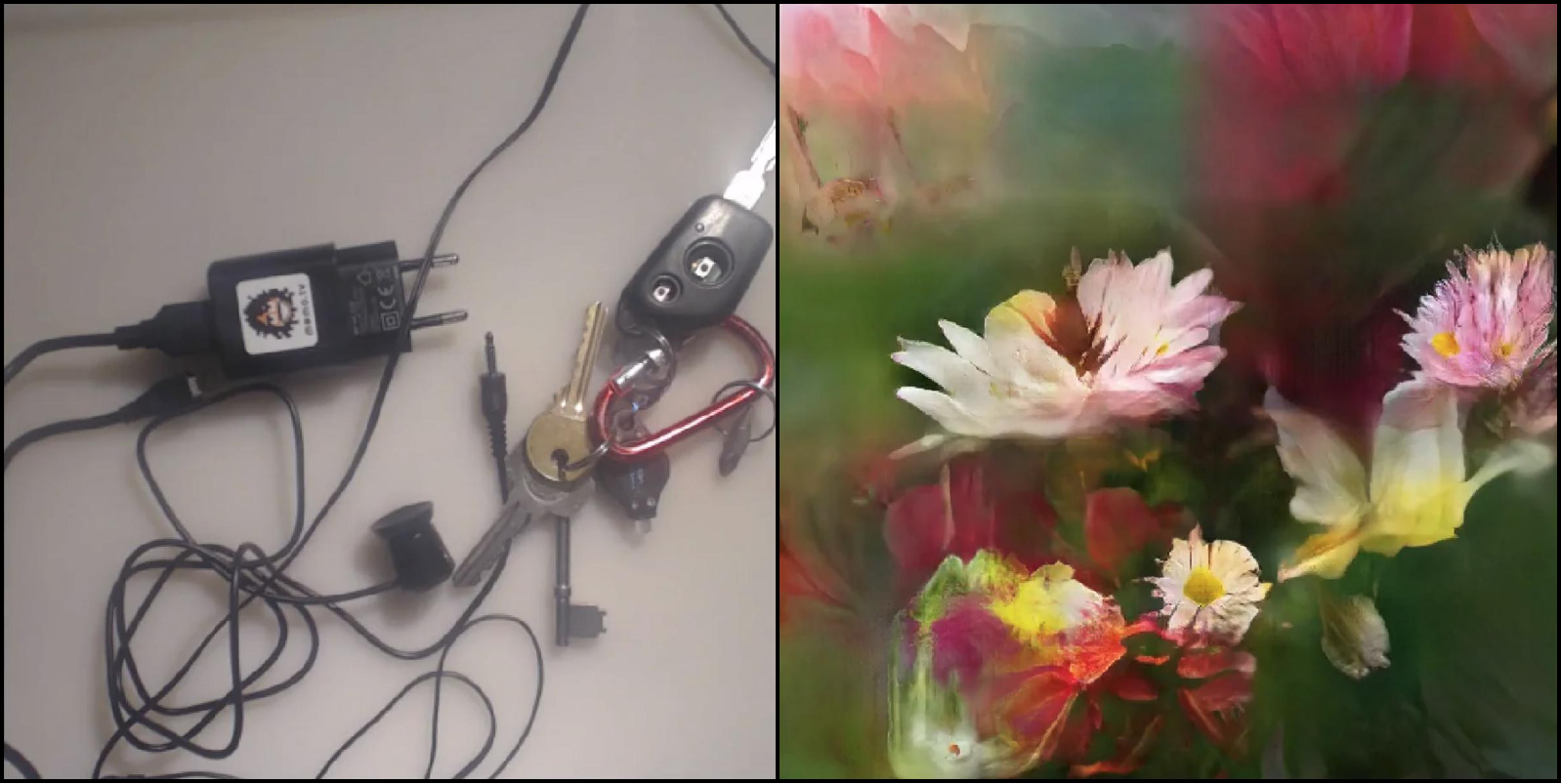}}
    \vspace{-0.1in}
    \caption{\emph{Learning to see: Gloomy Sunday}: An interactive artistic demo developed by Memo Akten~\cite{sunday} based on our {\tt pix2pix} codebase. Please click the image to play the video in a browser.}
    \label{sunday}
    \vspace{-0.2in}
\end{figure}

Since the initial release of the paper and our {\tt pix2pix} codebase, the Twitter community, including computer vision and graphics practitioners as well as visual artists, have successfully applied our framework to a variety of novel image-to-image translation tasks, far beyond the scope of the original paper.  Figure~\ref{twitter} and Figure~\ref{sunday}  show just a few examples from the \#pix2pix hashtag, including \emph{Background removal}, \emph{Palette generation}, \emph{Sketch $\rightarrow$ Portrait}, \emph{Sketch$\rightarrow$Pokemon}, \emph{"Do as I Do" pose transfer}, \emph{Learning to see: Gloomy Sunday}, as well as the bizarrely popular \#edges2cats and \#fotogenerator.
Note that these applications are creative projects, were not obtained in controlled, scientific conditions, and may rely on some modifications to the {\tt pix2pix} code we released. Nonetheless, they demonstrate the promise of our approach as a generic commodity tool for image-to-image translation problems.

\vspace{-0.06in}

\section{Conclusion}
The results in this paper suggest that conditional adversarial networks are a promising approach for many image-to-image translation tasks, especially those involving highly structured graphical outputs. These networks learn a loss adapted to the task and data at hand, which makes them applicable in a wide variety of settings.

\vspace{-0.06in}

\paragraph{Acknowledgments:} We thank Richard Zhang, Deepak Pathak, and Shubham Tulsiani for helpful discussions, Saining Xie for help with the HED edge detector, and the online community for exploring many applications and suggesting improvements. Thanks to Christopher Hesse, Memo Akten, Kaihu Chen, Jack Qiao, Mario Klingemann, Brannon Dorsey, Gerda Bosman, Ivy Tsai, and Yann LeCun for allowing the use of their creations in Figure~\ref{twitter} and Figure~\ref{sunday}. This work was supported in part by NSF SMA-1514512, NGA NURI, IARPA via Air Force Research Laboratory, Intel Corp, Berkeley Deep Drive, and hardware donations by Nvidia. J.-Y.Z. is supported by the Facebook Graduate Fellowship. Disclaimer: The views and conclusions contained herein are those of the authors and should not be interpreted as necessarily representing the official policies or endorsements, either expressed or implied, of IARPA, AFRL or the U.S. Government.


\clearpage{\thispagestyle{empty}\cleardoublepage}
{\small
\bibliographystyle{ieee}
\bibliography{main}
}

\clearpage

\section{Appendix}

\subsection{Network architectures}

We adapt our network architectures from those in \cite{radford2015unsupervised}. Code for the models is available at \texttt{https://github.com/phillipi/pix2pix}.

Let \texttt{Ck} denote a Convolution-BatchNorm-ReLU layer with k filters. \texttt{CDk} denotes a Convolution-BatchNorm-Dropout-ReLU layer with a dropout rate of $50\%$. All convolutions are $4 \times 4$ spatial filters applied with stride 2. Convolutions in the encoder, and in the discriminator, downsample by a factor of 2, whereas in the decoder they upsample by a factor of 2.

\subsubsection{Generator architectures}

The encoder-decoder architecture consists of:\\
{\bf encoder:}\\
\texttt{C64-C128-C256-C512-C512-C512-C512-C512}
{\bf decoder:}\\
\texttt{CD512-CD512-CD512-C512-C256-C128-C64}

After the last layer in the decoder, a convolution is applied to map to the number of output channels (3 in general, except in colorization, where it is 2), followed by a Tanh function. As an exception to the above notation, BatchNorm is not applied to the first \texttt{C64} layer in the encoder. All ReLUs in the encoder are leaky, with slope 0.2, while ReLUs in the decoder are not leaky.

The U-Net architecture is identical except with skip connections between each layer $i$ in the encoder and layer $n-i$ in the decoder, where $n$ is the total number of layers. The skip connections concatenate activations from layer $i$ to layer $n-i$. This changes the number of channels in the decoder:

{\bf U-Net decoder:}\\
\texttt{CD512-CD1024-CD1024-C1024-C1024-C512\\-C256-C128}

\subsubsection{Discriminator architectures}

The {\bf $70 \times 70$ discriminator} architecture is:\\
\texttt{C64-C128-C256-C512}

After the last layer, a convolution is applied to map to a 1-dimensional output, followed by a Sigmoid function. As an exception to the above notation, BatchNorm is not applied to the first \texttt{C64} layer. All ReLUs are leaky, with slope 0.2.

All other discriminators follow the same basic architecture, with depth varied to modify the receptive field size:

{\bf $1 \times 1$ discriminator:} \\\texttt{C64-C128} (note, in this special case, all convolutions are $1 \times 1$ spatial filters)\\
{\bf $16 \times 16$ discriminator:} \\\texttt{C64-C128}\\
{\bf $286 \times 286$ discriminator:} \\\texttt{C64-C128-C256-C512-C512-C512}\\

\subsection{Training details}

Random jitter was applied by resizing the $256 \times 256$ input images to $286 \times 286$ and then randomly cropping back to size $256 \times 256$.

All networks were trained from scratch. Weights were initialized from a Gaussian distribution with mean 0 and standard deviation 0.02.

{\bf Cityscapes labels$\rightarrow$photo} 2975 training images from the Cityscapes training set \cite{Cordts2016Cityscapes}, trained for 200 epochs, with random jitter and mirroring. We used the Cityscapes validation set for testing. To compare the U-net against an encoder-decoder, we used a batch size of 10, whereas for the objective function experiments we used batch size 1. We find that batch size 1 produces better results for the U-net, but is inappropriate for the encoder-decoder. This is because we apply batchnorm on all layers of our network, and for batch size 1 this operation zeros the activations on the bottleneck layer. The U-net can skip over the bottleneck, but the encoder-decoder cannot, and so the encoder-decoder requires a batch size greater than 1. Note, an alternative strategy is to remove batchnorm from the bottleneck layer. See errata for more details.

{\bf Architectural labels$\rightarrow$photo} 400 training images from \cite{Tylecek13}, trained for 200 epochs, batch size 1, with random jitter and mirroring. Data were split into train and test randomly.

{\bf Maps$\leftrightarrow$aerial photograph} 1096 training images scraped from Google Maps, trained for 200 epochs, batch size 1, with random jitter and mirroring. Images were sampled from in and around New York City. Data were then split into train and test about the median latitude of the sampling region (with a buffer region added to ensure that no training pixel appeared in the test set).

{\bf BW$\rightarrow$color} 1.2 million training images (Imagenet training set \cite{russakovsky2015imagenet}), trained for $\sim 6$ epochs, batch size 4, with only mirroring, no random jitter. Tested on subset of Imagenet val set, following protocol of \cite{zhang2016colorful} and \cite{larsson2016learning}.

{\bf Edges$\rightarrow$shoes} 50k training images from UT Zappos50K dataset \cite{yu2014fine} trained for 15 epochs, batch size 4.  Data were split into train and test randomly.

{\bf Edges$\rightarrow$Handbag} 137K Amazon Handbag images from \cite{zhu2016generative}, trained for 15 epochs, batch size 4. Data were split into train and test randomly.

{\bf Day$\rightarrow$night} 17823 training images extracted from 91 webcams, from \cite{laffont2014transient} trained for 17 epochs, batch size 4, with random jitter and mirroring. We use 91 webcams as training, and 10 webcams for test.

{\bf Thermal$\rightarrow$color photos} 36609 training images from set 00--05 of \cite{hwang2015multispectral}, trained for 10 epochs, batch size 4. Images from set 06-11 are used for testing.

{\bf Photo with missing pixels$\rightarrow$inpainted photo} 14900 training images from \cite{doersch2012makes}, trained for 25 epochs, batch size 4, and tested on 100 held out images following the split of \cite{pathak2016context}.

\subsection{Errata}
For all experiments reported in this paper with batch size 1, the activations of the bottleneck layer are zeroed by the batchnorm operation, effectively making the innermost layer skipped. This issue can be fixed by removing batchnorm from this layer, as has been done in the public code. We observe little difference with this change and therefore leave the experiments as is in the paper.

\subsection{Change log}
{\bf arXiv v2} Reran generator architecture comparisons (Section \ref{analysis_of_gen_arch}) with batch size equal to 10 rather than 1, so that bottleneck layer is not zeroed (see Errata). Reran FCN-scores with minor details cleaned up (results saved losslessly as pngs, removed unecessary downsampling). FCN-scores computed using scripts at \texttt{https://github.com/phillipi/pix2pix/tree/\\master/scripts/eval\_cityscapes}, commit d7e7b8b. Updated several figures and text. Added additional results on thermal$\rightarrow$color photos and inpainting, as well as community contributions.

{\bf arXiv v3} Added additional results on community contributions. Fixed minor typos.
\end{document}